%% file: main.tex
\documentclass{article}
\PassOptionsToPackage{numbers, compress}{natbib}

\usepackage[preprint]{neurips_2026}

\usepackage[utf8]{inputenc} 
\usepackage[T1]{fontenc}    
\usepackage{hyperref}       
\usepackage{url}            
\usepackage{booktabs}       
\usepackage{amsfonts}       
\usepackage{nicefrac}       
\usepackage{microtype}      
\usepackage{xcolor}         

\usepackage[table,x11names]{xcolor}  
\usepackage{tcolorbox}
\usepackage[normalem]{ulem}

\usepackage{graphicx}

\usepackage{amsmath}
\usepackage{amssymb}
\usepackage{mathtools}
\usepackage{amsthm}
\usepackage{multirow}
\usepackage{enumitem}
\usepackage{mathabx,graphicx}
\usepackage{algpseudocode}
\usepackage{algorithm}
\usepackage{wrapfig}
\usepackage[normalem]{ulem}
\usepackage{amssymb}
\usepackage{pifont}

\usepackage{caption}
\usepackage{xspace}

\usepackage{amssymb}
\usepackage{pifont}

\newcommand{\mypara}[1]{\noindent \textbf{#1}}

\newcommand{\worldwideweb}{\raisebox{-1.5pt}{\includegraphics[height=1.05em]{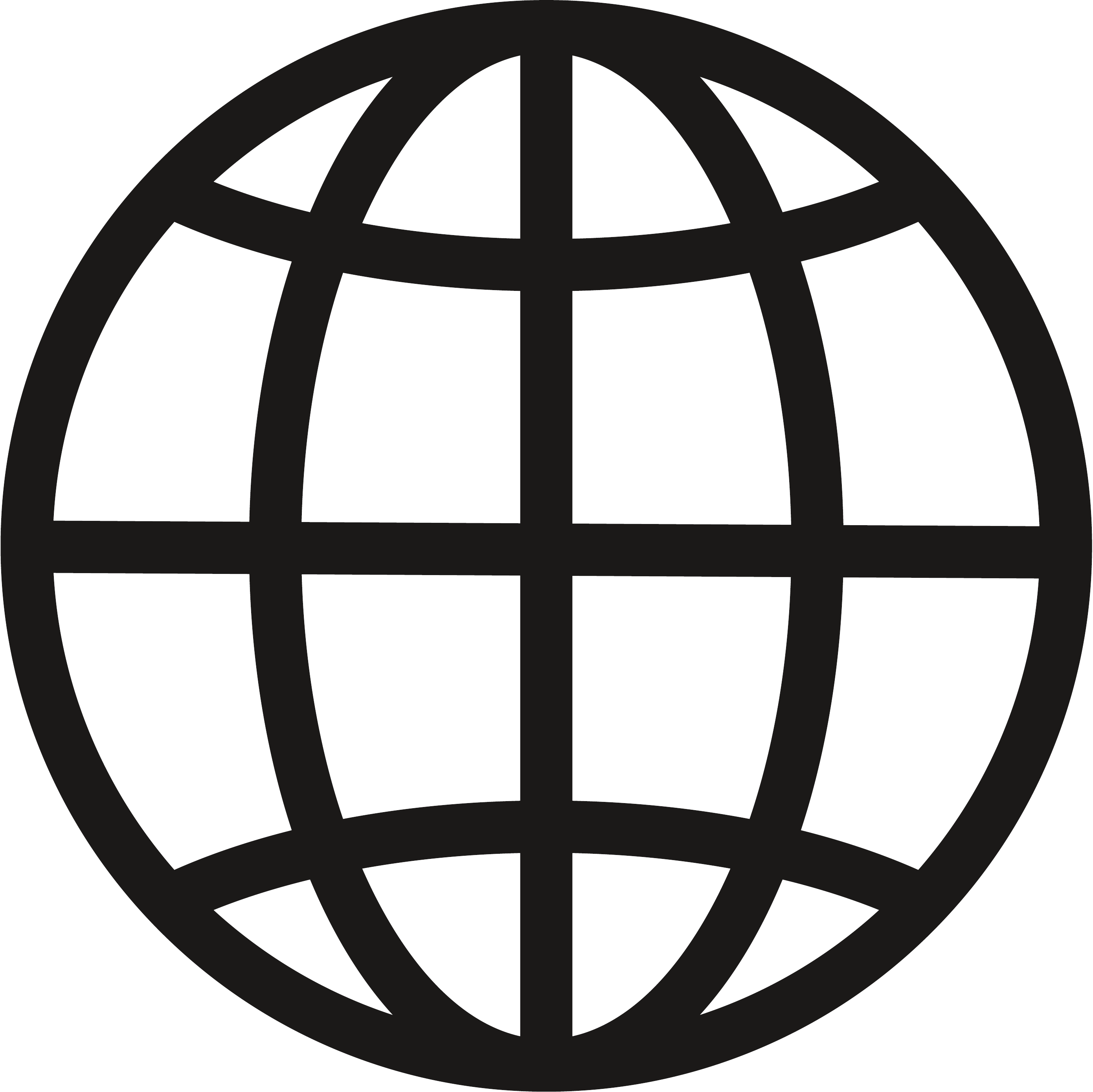}}\xspace}

\title{Ask When It Pays: Cost-Aware Open-Ended Interaction for Instance Goal Navigation}

\author{
Xunyi Zhao$^{1,2}$\thanks{Equal Contribution}
\quad Sihao Lin$^{1,2*}$
\quad Gengze Zhou$^{1}$ 
\quad Zerui Li$^{1}$ \\
\quad \textbf{Shijie Li}$^{3}$ 
\quad \textbf{Wei Tao} $^{4}$ 
\quad \textbf{Jiajun Liu}$^{2,5}$
\quad \textbf{Qi Wu}$^{1,2}$\thanks{Corresponding author}\\
$^1$Adelaide University
$^2$Responsible AI Research Centre, Australian Institute for Machine Learning \\
$^3$Institute for Infocomm Research (I2R), A*STAR
$^4$iMotion 
$^5$CSIRO Data61 \\
{\worldwideweb \href{https://billzhao1030.github.io/IVLN}{{\text{Project Website}}}}
}

\begin{document}

\maketitle

\input{sec/0_abstract}
\input{sec/1_intro}
\input{sec/2_related}
\input{sec/3_method}
\input{sec/4_result}
\input{sec/5_conclusion}


{
    \small
    \bibliographystyle{plainnat}
    \bibliography{main}
}


\appendix
\input{sec/6_appendix}


\newpage

\end{document}

%% file: sec/0_abstract.tex
\begin{abstract}

Instance Goal Navigation (IGN) requires an embodied agent to find a \emph{specific} object instance among distractors from an underspecified natural-language description. Such ambiguity often cannot be resolved from perception and language alone, making interaction with an oracle a natural mechanism for disambiguation. Prior interactive methods allow oracle queries but treat lightweight clarification and route-level guidance alike, letting agents boost success rate through repeated high-information questions rather than by resolving the underlying ambiguity efficiently. 
We recast interactive IGN as a \emph{cost-sensitive uncertainty-reduction problem}, where the agent should ask the question whose answer provides the largest reduction in navigation uncertainty relative to its penalty.
To this end, we apply an information-gain analysis on existing navigation corpora to identify which cues reduce navigation uncertainty, yielding a compact set of question types and data-derived weights.
However, existing interactive navigation benchmarks do not model the cost of different question types or evaluate how efficiently agents use interaction, making them unsuitable for studying cost-sensitive interaction. 
Based on this taxonomy, we construct a benchmark for diagnosing interaction behavior and efficiency, together with a Weighted Success Rate metric that penalizes each query by its derived cost.
We further propose a zero-shot MLLM navigator that selectively queries at each decision step only when the expected uncertainty reduction justifies the interaction cost.
\end{abstract}

%% file: sec/1_intro.tex
\section{Introduction}
\label{sec:intro}

A household robot asked to \emph{``bring me the cup''} may face uncertain room boundaries, several same-category instances, and routes that reach different but seemingly valid candidates. This is \textbf{Instance Goal Navigation (IGN)}. Unlike Vision-and-Language Navigation (VLN)~\citep{anderson2018r2r}, where language often specifies a path, or ObjectNav~\citep{batra2020objectnav}, where any object from a target category may suffice, IGN requires resolving instance-level ambiguity from a description in a scene with distractors. The objective can remain ill-posed from perception and the coarse instruction alone, because room context, object attributes, spatial relations, and partial routes may each support multiple targets~\citep{yin2025hypernav, sun2026view, zhong2026spatial, li2025ground}. Interaction with a knowledgeable oracle is therefore a natural way to make the task well specified. This raises three coupled design questions: \emph{when} to ask, \emph{what} to ask, and \emph{how the agent uses the answer}. We review the dialog-navigation lineage in \S\ref{sec:related} and turn next to its open problems.

Existing interactive navigation work~\citep{thomason2020cvdn, nguyen2019hanna, nguyen2019vnla, chi2020justask, gao2022dialfred, taioli2025coin, han2025dialnav, huang2025vlln} has shown that agents can query an oracle and use the answers to navigate, including in zero-shot MLLM settings. Yet most formulations still leave two issues under-specified. The space of possible questions is hand-designed, and the cost of asking is absent or treated uniformly. This matters because a target-confirmation question and a route-guidance question provide very different amounts of oracle information. Under cost-agnostic evaluation, an agent can improve Success Rate (SR) by repeatedly asking high-information route questions, while raw SR cannot separate efficient ambiguity resolution from heavy oracle use.

We recast interactive IGN as a cost-sensitive uncertainty-reduction problem. The agent should ask the question whose answer is expected to reduce the most navigation uncertainty relative to its query cost. The missing link is explicit knowledge of which linguistic cues reduce uncertainty and how queries for those cues should be priced~\cite{choudhury2025bed}. We build this link by applying an information-gain-inspired analysis to heterogeneous human-authored navigation corpora. The mined cues are collapsed into four question types covering appearance, location or region, direction or route, and target verification, with locked per-type costs derived from corpus statistics. Route-level questions remain available because they are useful, but their higher information value is reflected in evaluation.

To study this formulation, we need episodes where ambiguity is explicit, controllable, and connected to grounded oracle answers. We therefore construct an interactive IGN benchmark in Isaac Sim with object-instance annotations, room metadata, controlled same-category distractors, and type-specific oracle answers grounded to the target instance~\cite{Mittal_Isaac_Lab_-_2025, makoviychuk2021isaac}. Each episode gives the agent a coarse goal such as ``Find a laptop'' and requires it to identify the intended instance without a human route trajectory~\cite{roman2020rmm, thomason2020cvdn,zhu2021soon, qi2020reverie, nguyen2019hanna}. Difficulty is assigned before evaluation from distractor count, same-room distractors, contextual ambiguity, and initial path distance, then episodes are sampled into balanced easy, medium, and hard splits. We evaluate with Weighted Success Rate, which discounts successful navigation by the accumulated query costs while reducing to SR when no question is asked~\cite{anderson2018r2r, krantz2020beyond, lin2025vlnverse, chang2017matterport3d, ramakrishnan2021hm3d, habitat19iccv, anderson2018evaluation}. Together, the controlled difficulty factors, typed question traces, and weighted metric let us diagnose whether failures arise from target identity, spatial uncertainty, distractor verification, or inefficient oracle use.

We then instantiate this setting with a modular zero-shot MLLM navigator, \textbf{TANDEM}, via \textbf{T}wo-st\textbf{A}ge \textbf{N}avigation with \textbf{D}isentangl\textbf{E}d planning and \textbf{M}etric grounding. The design decouples navigation planning from 3D grounding and metric execution. A Navigator MLLM reasons over the current observation, symbolic spatial memory, and accumulated oracle facts, maintaining context about explored spaces and object evidence. Instead of asking the MLLM to manipulate metric 3D state directly, a Grounder maps the Navigator's plan to a local waypoint, and a deterministic executor handles motion and collision. The Navigator also decides whether to ask a cost-aware question when important cues remain unresolved. To prevent interaction from replacing navigation, the oracle provides natural-language spatial cues but never metric coordinates, compass angles, or direct action commands.

Across MLLM backbones and simulator settings, TANDEM improves Weighted SR over no-interaction and uniform-cost interactive comparisons, with the largest gains on hard episodes involving many distractors, ambiguous rooms, or visually similar objects. The diagnostic results show that useful interaction is structured rather than uniform. Appearance and region questions reduce early target uncertainty, route questions resolve spatial uncertainty at ambiguous branches, and verification questions prevent stopping on distractors. 
Our contributions are:
\begin{itemize}
    \item We formulate interactive IGN as cost-sensitive uncertainty reduction and derive a four-type question taxonomy with data-derived penalties from human navigation corpora.
    \item We build a controllable instance-level IGN benchmark with grounded oracle answers, balanced difficulty splits, and Weighted Success Rate for diagnosing interaction efficiency.
    \item We provide TANDEM as a modular zero-shot MLLM baseline and show that cost-aware interaction, spatial memory, and disentangled grounding improve navigation performance.
\end{itemize}

%% file: sec/2_related.tex
\section{Related Work}
\label{sec:related}

\textbf{Interactive Embodied Navigation.}
Existing interactive vision-and-language navigation has explored several forms of language-based assistance, but the role and cost of different question types remain under-specified. Early work often frames interaction as asking for the next navigation decision, where the oracle returns the optimal action or a paraphrased route segment~\citep{thomason2020cvdn,nguyen2019vnla,chi2020justask,qiao2023march,gao2022dialfred}. Even when asking is uncertainty-gated, the answer still exposes trajectory-level guidance~\citep{nguyen2019vnla,chi2020justask, ma2019self, ma2019regretful, padmakumar2022teach}. Later methods provide less prescriptive spatial, visual, or object-level cues~\citep{nguyen2019hanna}, with recent extensions annotating attribute, disambiguation, and spatial information~\citep{han2025dialnav, banerjee2021robotslang}. However, these methods still depend on large annotated dialog pools and trained navigators or oracles. Zero-shot prompting reduces this burden, but typically uses hand-fixed interaction structures: COIN allows only object-attribute or region questions and forbids route questions~\citep{taioli2025coin}, while a concurrent interactive IGN benchmark fixes description, route, and confirmation questions by episode phase~\citep{huang2025vlln}. Overall, prior work shows that interaction improves navigation, but does not connect question categories to their uncertainty-reduction value or charge agents for different oracle information. We address this gap with a cost-aware formulation that evaluates interaction by both what it resolves and what it costs.

\textbf{Zero-shot MLLM for Embodied Navigation.}
Zero-shot navigation with large vision-language models has become a popular alternative to task-specific training, with the agent translating observations and history into language~\cite{zhou2024navgpt, zhou2025navgpt2, pan2023langnav, long2023discuss, long2024instructnav, lin2025vlnverse} and emitting waypoints~\cite{qiao2024open, shi2025smartway, zhang2026spatialnav}, map nodes~\cite{chen2024mapgpt, zhang2025mapnav}, or low-level actions~\cite{zhang2024navid, cheng2024navila, zhang2024uni,wei2025ground}. Two limitations of this approach are now well documented. First, although MLLMs achieve strong visual understanding, they remain limited in 3D spatial reasoning, in tracking history across multiple decision steps, and in forecasting the outcome of a chosen action~\cite{qiao2025navbench,chen2024spatialvlm, zhao2025vln}. Second, the action interface itself introduces additional failure modes. Waypoint-based and graph-based navigators are bounded by the coverage and accuracy of an external proposal predictor, whereas end-to-end variants require the language model itself to produce metric quantities such as turn angles and step distances, for which it lacks reliable physical grounding and often fails~\cite{krantz2020beyond, krantz2021waypoint,hong2022bridging, an2023etpnav, an2022bevbert, zhang2026spatialant, wang2023gridmm, wang2025rethinking}. Together, these limitations motivate keeping the language model on high-level reasoning and pairing it with an executor that can faithfully realize its plan, which is the gap our agent closes.

%% file: sec/3_method.tex
\section{Method}
\label{sec:method}

We model interactive instance-goal navigation as cost-sensitive uncertainty reduction, where interaction acquires typed cues that reduce target or spatial ambiguity. The three components form a single protocol. The question taxonomy defines what information can be requested and how each request is priced (\S\ref{sec:method:uncertainty}). The benchmark instantiates this taxonomy with grounded answer sources for each question type (\S\ref{sec:method:benchmark}). TANDEM then operates under the same protocol, deciding when to ask, how to store the answer, and how to ground the resulting plan into motion (\S\ref{sec:method:agent}).

\subsection{Question Utility Modeling}
\label{sec:method:uncertainty}

We derive the question taxonomy and its interaction costs from an information-gain-inspired uncertainty model. Useful questions should reduce navigation uncertainty, and their costs should reflect how informative the corresponding cues are in human navigation language. Let $S$ denote the hidden target state. The intrinsic value of a candidate question $q$ with answer $a$ is its expected information gain (EIG)~\cite{lindley1972}:
\begin{equation}
    \mathrm{EIG}(q) \;=\; \mathrm{H}\!\left[p_t(S)\right] - \mathbb{E}_{a \sim p(a\mid q)}\!\left[\mathrm{H}\!\left[p_t(S \mid q, a)\right]\right].
    \label{eq:eig}
\end{equation}
Eq.~\eqref{eq:eig} motivates cost-sensitive interaction: the agent should ask questions that reduce $\mathrm{H}[p_t(S)]$ substantially relative to their cost. Directly computing this quantity is impractical because the posterior $p_t(S\mid q,a)$ over real targets in real scenes is unobservable. We therefore use heterogeneous human-authored navigation corpora as an empirical proxy for which cues typically reduce target and spatial ambiguity. Rather than estimating exact posterior information gain, this mining step provides a reproducible, data-derived prior for assigning question costs.

We canonicalize records from R2R~\cite{anderson2018r2r}, REVERIE~\cite{qi2020reverie}, RxR~\cite{anderson2020rxr}, CVDN~\cite{thomason2020cvdn}, and SOON~\cite{zhu2021soon}, covering route instructions, dialog histories, and object-grounded descriptions. A fixed-schema Qwen3.5-4B scorer labels each record with a closed set of prior cues, evidence spans, an importance score, an uncertainty-reduction label, and a rank. We then group the mined cues by answer semantics into four interaction-level question types, so the taxonomy is induced from the uncertainty-mining ontology while each type remains tied to the ambiguity its answer is meant to reduce.

\noindent\textbf{From cue utility to per-type cost.}
For each cue $c$, we compute a composite utility score from three observable mining statistics:
\begin{equation}
    U_c \;=\; \lambda_1\,\bar{s}_c \;+\; \lambda_2\,h_c \;+\; \lambda_3\,r_c,
    \label{eq:uc}
\end{equation}
where $\bar{s}_c$ is the average scorer-assigned importance, $h_c$ is the fraction of cue mentions marked as high-gain, and $r_c$ is the fraction of records in which the cue is ranked first. Cue utilities are mention-weighted within each question type and rescaled into locked per-type costs $w_t$ (Appendix~\ref{app:uncertainty}). More informative question types receive larger penalties because they expose stronger oracle assistance and should be discounted more heavily when measuring autonomous navigation ability. Table~\ref{tab:method:qtypes} summarizes the resulting taxonomy, the primary uncertainty reduced by each answer type, and the locked cost used by the benchmark metric defined next.

\begin{table}[h]
    \centering
    \small
    \setlength{\tabcolsep}{4pt}
    \renewcommand{\arraystretch}{1.1}
\vspace{-2pt}
    \caption{Question types induced by uncertainty mining and their per-type cost coefficient $w_t$.}
    \label{tab:method:qtypes}
    \begin{minipage}{0.9\linewidth}
    \centering
    \begin{tabular}{@{}>{\bfseries}p{0.30\linewidth}p{0.52\linewidth}c@{}}
        \toprule
        Question type & Primary uncertainty reduced & Penalty $w_t$ \\
        \midrule
        Type~1 appearance & Target identity among same-category instances & 0.182 \\
        Type~2 region & Room or region to search & 0.162 \\
        Type~3 direction/route & Branch, corridor, or layout direction to follow & 0.240 \\
        Type~4 confirmation & Whether a visible candidate is the target & 0.103 \\
        \bottomrule
    \end{tabular}
\vspace{-4pt}
    \end{minipage}
\end{table}

\subsection{Benchmark Construction}
\label{sec:method:benchmark}

To compare with prior baselines, we first evaluated agents on existing dialog and instruction-following benchmarks~\cite{thomason2020cvdn, qi2020reverie, zhu2021soon}. Cost-sensitive interactive IGN also requires state that those datasets do not expose at runtime: persistent object identities, room membership, controllable same-category distractors, and typed answer sources tied to the target instance. We therefore construct an instance-level benchmark in Isaac~Sim using USD scenes~\cite{usd}, whose metadata gives a consistent object, room, pose, and relation graph.

Each episode gives a coarse goal such as ``Find a laptop'' and asks the agent to identify the intended instance among same-category distractors, without a human route trajectory. The released record contains the scene id, difficulty label, instruction, start pose, target category and instance id, goal room and region, reference-view id, natural-language target description, and oracle metadata. Construction-only fields such as distractor lists, target boxes, and structured attributes are retained for validation but hidden when they would leak the answer.

For each scene, we build an instance inventory and room-object graph, keep manually verified discrete object categories, cap extremely frequent categories, enumerate same-category distractors, and sample reachable starts. For each target, we use a target-facing reference view and ask Qwen3.5-4B to verify visibility, then generate a compact appearance-and-relation description without room names. Post-generation checks remove wrong-category references, room leakage, empty descriptions, and descriptions that do not distinguish the target from its distractors. These fields provide Type~1 and Type~2 answers, while Type~3 and Type~4 are generated from route renders, occupancy maps, and visibility checks as described in \S\ref{sec:method:agent}.

\begin{figure}[t]
    \centering
    \includegraphics[width=0.9\linewidth]{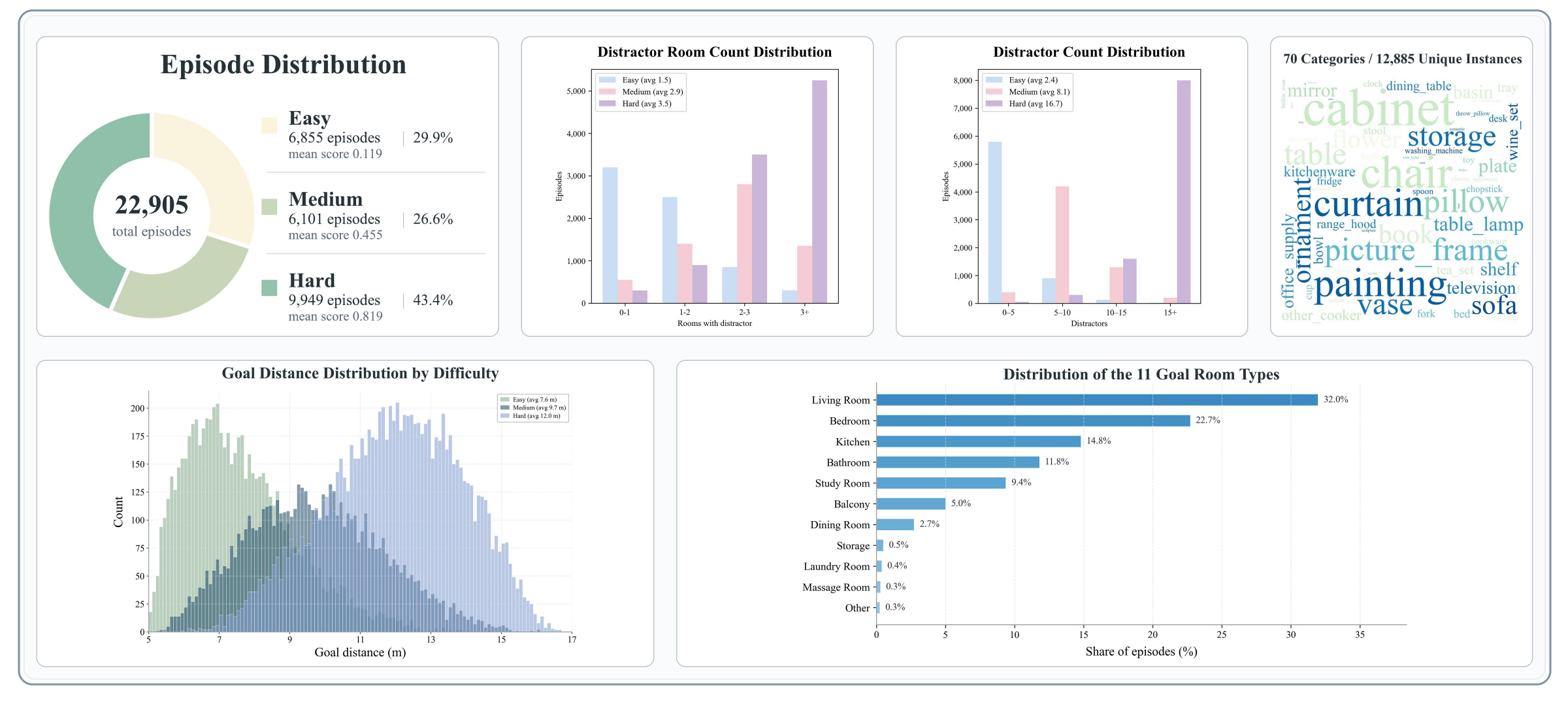}
\vspace{-2pt}
    \caption{\textbf{Benchmark statistics.} Overview of the dataset composition, including episode distribution by difficulty, distractor room and instance counts, target object categories, goal-distance distribution, and goal-room distribution.}
    \label{fig:method:bench-stats}
\vspace{-10pt}
\end{figure}

Each episode receives a deterministic difficulty score from distractor count, same-room distractors, contextual ambiguity, and the initial path distance. We bin episodes into easy, medium, and hard, then sample a 30:40:30 evaluation split from an eligible corpus of $22{,}905$ episodes across 262 scenes, 70 target categories, and 11 normalized goal-room labels. The evaluation subset used in \S\ref{sec:exp} contains 500 episodes. Figure~\ref{fig:method:bench-stats} summarizes the full-benchmark statistics, while Appendix~\ref{app:benchmark} details the filtering rules, difficulty formula, and geometry validation checks.

\noindent\textbf{Weighted Success Rate.}
Standard SR ignores how much oracle information was consumed. We therefore charge each question by its per-type cost $w_t$ from Eq.~\eqref{eq:uc} and define the episode score:
\begin{equation}
    \mathrm{Weighted\,SR}_e(\tau)
    =
    \mathbf{1}\{d_e \le \tau\}
    \exp\!\left(-\sum_{i=1}^{N_e} w_{\mathrm{type}(q_{e,i})}\right)
    \label{eq:wsr}
\end{equation}

where $d_e$ is the final distance to the target, $\tau$ is the success threshold, $N_e$ is the number of questions in episode $e$, and $\mathrm{type}(q_{e,i})$ is the type of the $i$-th question. The discount lies in $[0,1]$, repeated questions are charged repeatedly, and successful no-question episodes reduce to plain success. Failed episodes receive zero. The dataset-level average is given in Appendix~\ref{app:uncertainty}.

\subsection{TANDEM: Two-stage Navigation with Disentangled Planning and Metric Grounding}
\label{sec:method:agent}

TANDEM instantiates the benchmark protocol as a stateful zero-shot MLLM navigator. Each step has exactly one high-level outcome. The agent either asks one typed question, moves toward a grounded local subgoal, or stops with target-confirmation evidence.

\noindent\textbf{Two-stage planning and grounding.}
At step $t$, following the compass-aligned observation design of SpatialNav~\cite{zhang2026spatialnav}, the simulator renders an 8-view panorama $O_t=\{o_t^k\}_{k=0}^{7}$, where each $o_t^k$ is a $90^\circ$ Field-of-View RGB view and adjacent view centers are $45^\circ$ apart. The Planner does not receive a metric map. Instead, it observes $\mathcal{X}_t=(O_t,g,H_t,R_t,F_t,M_t)$, where $g$ is the coarse goal, $H_t$ is the movement history, $R_t$ is the remaining question budget, $F_t$ is the Fact Base of oracle-confirmed information, and $M_t$ is a structured Spatial Memory. It then produces one semantic decision,
\begin{equation}
    z_t=\Pi_\theta(\mathcal{X}_t), \quad
    z_t \in
    \{\textsc{ASK}(\ell_t,q_t),\,
    \textsc{MOVE}(k_t,u_t,\rho_t,\Delta M_t),\,
    \textsc{STOP}(\rho_t)\}.
    \label{eq:tandem-planner}
\end{equation}
Here $\ell_t\in\{1,2,3,4\}$ is the question type, $q_t$ is the question text, $k_t$ selects one panorama view, $u_t$ is a place-based local subgoal, $\rho_t$ records the purpose and high-level reasoning, and $\Delta M_t$ is the Planner's belief update. A move decision is deliberately semantic rather than metric. The Planner chooses a direction and describes the intended traversable place, while the Grounder receives only the selected $90^\circ$ view, the local subgoal, and the purpose, then selects one cell on an $8\times8$ grid as the concrete local target. The executor back-projects this cell through the camera model and floor-plane geometry into a short world-frame waypoint, then applies deterministic local control. Before committing to motion, it casts forward rays at multiple heights. If any ray detects an obstacle within $0.6$\,m, the executor turns toward the nearest collision-free neighboring heading and writes the blocked direction into Spatial Memory.

\begin{figure}[t]
    \centering
    \includegraphics[width=0.95\linewidth]{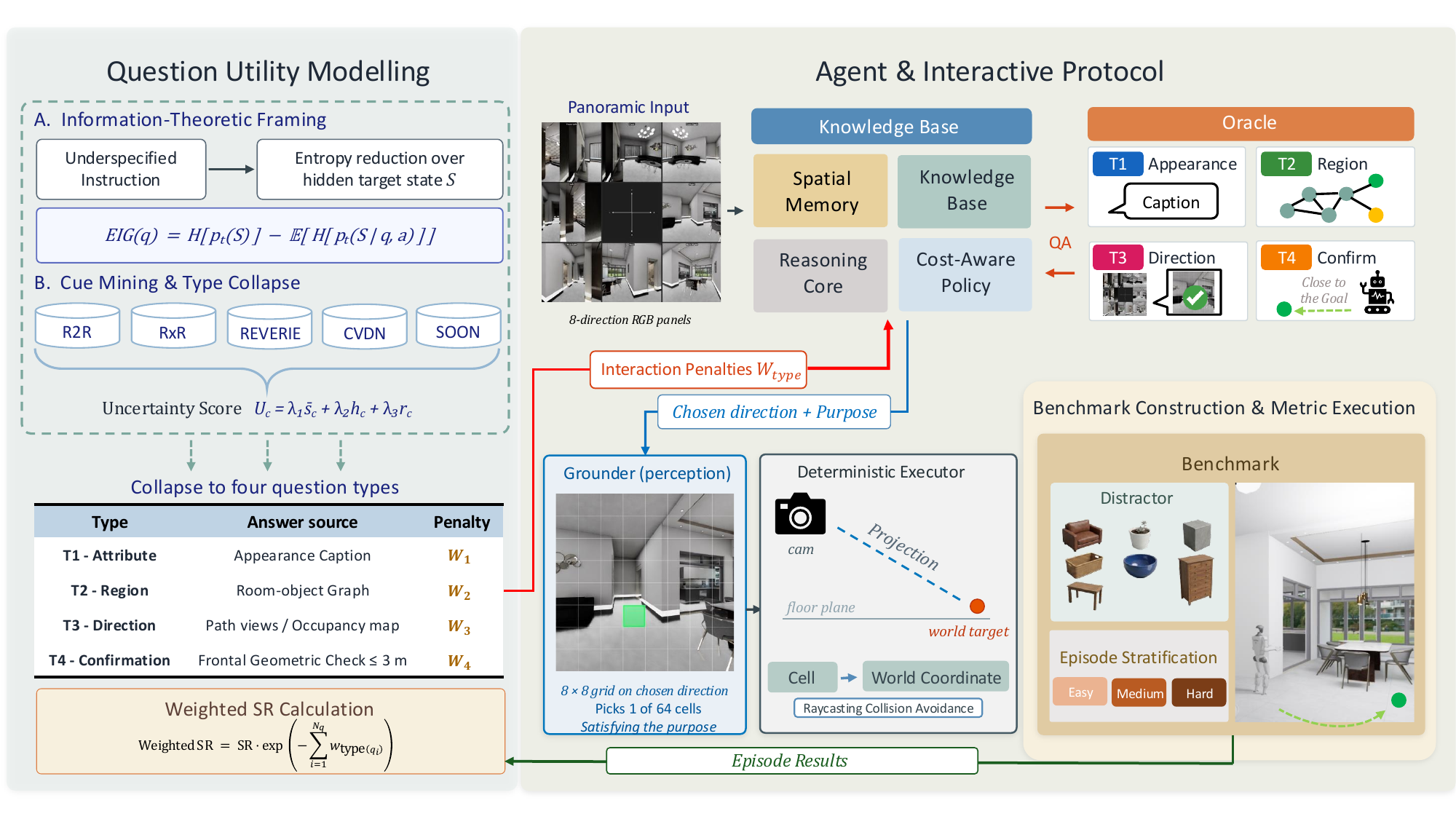}
\vspace{-2pt}
    \caption{TANDEM decomposes interactive instance image-goal navigation into two coupled stages. \textbf{(Left) }Question Utility Modelling\textbf{ (\S~\ref{sec:method:uncertainty}).}  \textbf{(Right) }Agent and Interactive Protocol\textbf{ (\S~\ref{sec:method:agent})}}
    \label{fig:method:arch}
\vspace{-12pt}
\end{figure}

After each step, the controller merges the Planner's belief update with newly observed openings, candidate objects, oracle facts, and execution feedback. Spatial Memory is thus a language-level record of visited regions, unexplored openings, blocked moves, route hints, and candidate checks, not an occupancy map exposed to the MLLM. Because pose, grounded waypoints, and active spatial hints share the same world frame, the controller can recompute bearing and distance after detours rather than losing the intended route after a turn. This separation keeps the MLLM's role qualitative and makes metric quantities, collision handling, and pose updates auditable. Figure~\ref{fig:method:arch} shows the full flow, and Appendix~\ref{app:agent} gives the prompt schema, observation ablation, and local-control details.

\noindent\textbf{Cost-aware interaction and typed oracle.}
Cost awareness comes from the same stateful interface rather than from a separate learned cost model. At every Planner call, the agent observes the remaining question budget, the relative penalties of the four question types, the Fact Base, and Spatial Memory. If it asks, no movement is executed in that step. The controller sends the selected type and natural-language question to the Oracle, charges the corresponding penalty in Eq.~\eqref{eq:wsr}, stores the answer in the Fact Base, and exposes it to later Planner calls. Thus duplicate or low-value questions compete against information already resolved by earlier answers, observations, and failed moves.

The Oracle follows the same four-type interface used in the benchmark metric, so different answers reduce different uncertainties and incur the corresponding penalties. Type~1 answers use the sanitized target appearance description and cannot reveal room or route cues. Type~2 answers use the room-object graph and may name the room or region, but cannot add appearance details.

Type~3 answers are generated from two internal sources, a front-view clip rendered over the first $4$\,m of the shortest path from the current pose toward the target and occupancy-map metadata such as free-space connectivity, openings, and room adjacency. The Oracle summarizes these sources as a natural-language route or layout hint, but the Planner never receives the raw clip, occupancy map, camera poses, waypoint coordinates, path length, compass heading, step count, final target coordinate, or exact action sequence. Type~4 verifies a visible candidate using the current front view and geometric visibility checks, returning a positive answer only when the queried candidate is the target, visible in front, and within $3$\,m. Otherwise, it returns ``no'' or a short negative rationale. A positive Type~4 answer creates a follow-up state in which the Planner either stops if close enough or takes one closer move before stopping. Stop decisions are therefore tied to explicit confirmation or to a spatial hint that places the agent within the success threshold, rather than to the Grounder alone.

%% file: sec/4_result.tex
\section{Experiments}
\label{sec:exp}

\subsection{Experimental Setup}
\label{sec:exp:setup}

\mypara{Benchmarks and baselines.} We evaluate TANDEM on a fixed pool of 500 episodes sampled from the full benchmark in \S\ref{sec:method:benchmark}, including 150 easy, 200 medium, and 150 hard episodes, balanced across 11 goal-room categories. The same episode pool is used for all methods, ensuring paired comparisons at the episode level. We compare TANDEM with prior zero-shot agents on the proposed benchmark, including GTA~\cite{li2026gta}, MapGPT~\cite{chen2024mapgpt}, NavGPT~\cite{zhou2024navgpt}, and COIN~\cite{taioli2025coin}. All methods are adapted to a unified setup with the same simulator, observation interface, oracle protocol, and scoring rule, using Qwen3.5-8B~\cite{qwen2026qwen35} as the default backbone. To assess generalization, we further evaluate on CVDN~\cite{thomason2020cvdn}, SOON~\cite{zhu2021soon}, and REVERIE~\cite{qi2020reverie} under the VLN-MME~\cite{zhao2025vln} framework to compare with prior supervised instruction-following methods, including HAMT~\cite{chen2021hamt}, DUET~\cite{chen2022duet}, AutoVLN~\cite{chen2022hm3dlearning}, GOAT~\cite{wang2024vision}, ScaleVLN~\cite{wang2023scaling}, NaviLLM~\cite{zheng2023towards}, and SAME~\cite{zhou2025same}.



\mypara{Evaluation protocol.}
For a threshold $\tau$, $\mathrm{SR}@\tau$ measures success as reaching within $\tau$ meters of the target instance. The conventional VLN radius of $3$\,m is too permissive for instance-level navigation, as multiple distractors from the same category may fall within this range. We therefore adopt $\mathrm{SR}@1.5$\, as the primary metric, report $\mathrm{SR}@0.5$\, as a strict localization check, and retain $\mathrm{SR}@3$\, for comparability with prior work. In addition to $\mathrm{SR}/\mathrm{OSR}$, we report navigation error (NE), trajectory length (TL), per-type and total question counts ($Q_1,\ldots,Q_4$, $|Q|$), and the cost-aware Weighted~SR (Eq.~\eqref{eq:wsr}) evaluated at $\{1.5, 3\}$\,m.

\subsection{Main Results}
\label{sec:exp:main}

\input{sec/tab/agent_table}
\input{sec/tab/backbone_table}



\mypara{Comparison to zero-shot agents.}
To ensure a fair comparison, all methods adopt Qwen3.5-8B as the backbone. We equip prior methods with the same question taxonomy and evaluate them under two QA settings to assess the effectiveness of the proposed cost-aware interaction. First, agents are allowed to freely interact with the oracle at each step. Second, we impose the same cost-aware interaction constraint as in TANDEM. As shown in Table~\ref{tab:agent_results}, TANDEM achieves 35.3 SR@1.5 and 21.4 Weighted SR, consistently outperforming prior approaches.
For example, under the cost-aware setting, TANDEM surpasses the second-best method, GTA, by 9.3 points in SR@1.5. Interestingly, although naive interaction allows agents to ask more questions, it does not lead to clear performance gains and can even degrade performance as the interaction cost increases. This observation highlights that simply increasing the number of queries is insufficient. In contrast, the proposed cost-aware interaction effectively regulates query cost in interactive IGN, while achieving comparable or even superior performance to widely used naive interaction.

\input{sec/tab/simulator_result}

\mypara{Comparison on other benchmarks.} Beyond the proposed benchmark, Table~\ref{tab:simulator_compact} presents a compact comparison on CVDN, SOON, and REVERIE. These results are included as a transfer check as the three benchmarks differ substantially in instructions, splits, and evaluation protocols. Despite these differences, TANDEM consistently improves the held-out metrics, suggesting that its gains are not tied to the new benchmark alone.

\mypara{Ablating interactive modules.}
Table~\ref{tab:agent_results} presents a modular ablation study isolating the two key components in TANDEM for navigator-oracle interaction: Spatial Memory and QA. Removing Spatial Memory reduces SR@1.5 from 35.3 to 28.0 ($-7.3$), while removing QA leads to a larger drop to 20.0 ($-15.3$). When both components are removed, TANDEM degenerates to a naive zero-shot navigator, achieving only 14.0 SR@1.5. The QA-off setting corresponds to the traditional non-interactive IGN setup, where a full structured description is provided upfront. The substantial 15.3-point gap compared to full TANDEM indicates that even complete information, combined with structured memory, cannot substitute for targeted interaction. Interestingly, the spmem-off variant still outperforms the QA-off variant (28.0 vs.\ 20.0 SR@1.5), which lacks Type~3 (route/layout) questions. This suggests that coarse spatial guidance plays a central role in our interactive framework rather than serving as a minor enhancement. Finally, without QA interaction, TANDEM tends to terminate exploration prematurely (as reflected by the TL metric), indicating that interaction provides critical information to reduce navigation uncertainty and sustain effective exploration toward the IGN goal.



\mypara{Ablating backbones.}
We study the impact of backbone selection across representative MLLMs, including the Qwen family~\cite{qwen2026qwen35, bai2025qwen3vl}, InternVL series~\cite{wang2025internvl35}, Gemma~\cite{deepmind2026gemma4}, GPT-5.4~\cite{openai2026gpt54}, and Gemini3-Flash~\cite{google2025gemini3}. Across Navigator backbones, performance scales consistently with model strength. As shown in Table~\ref{tab:backbone_results}, GPT-5.4 achieves the best overall performance (41.6 SR@1.5, 25.0 WSR@1.5), while Qwen3.5-8B is the strongest open backbone. Models with smaller capacity, such as Qwen3VL-4B, Qwen3.5-4B, InternVL3.5-4B, and Gemma-e4B, follow the same descending trend on both metrics. This consistent ranking suggests that stronger Navigators improve performance through better navigation capability, rather than increased reliance on interaction. 

\subsection{The Aha Moment: Quantifying When and Where TANDEM Asks}
\label{sec:exp:dynamics}


Our benchmark enables a fine-grained analysis of how TANDEM uses interaction to reduce navigation uncertainty, rather than merely evaluating final success.
We study the temporal distribution of question types, the physical contexts in which Type~3 spatial questions arise, and a proxy for the reduction in search space following Type~3 responses, offering insight into the role of interaction in navigation.

\begin{figure}[th]
    \centering
    \includegraphics[width=0.9\linewidth]{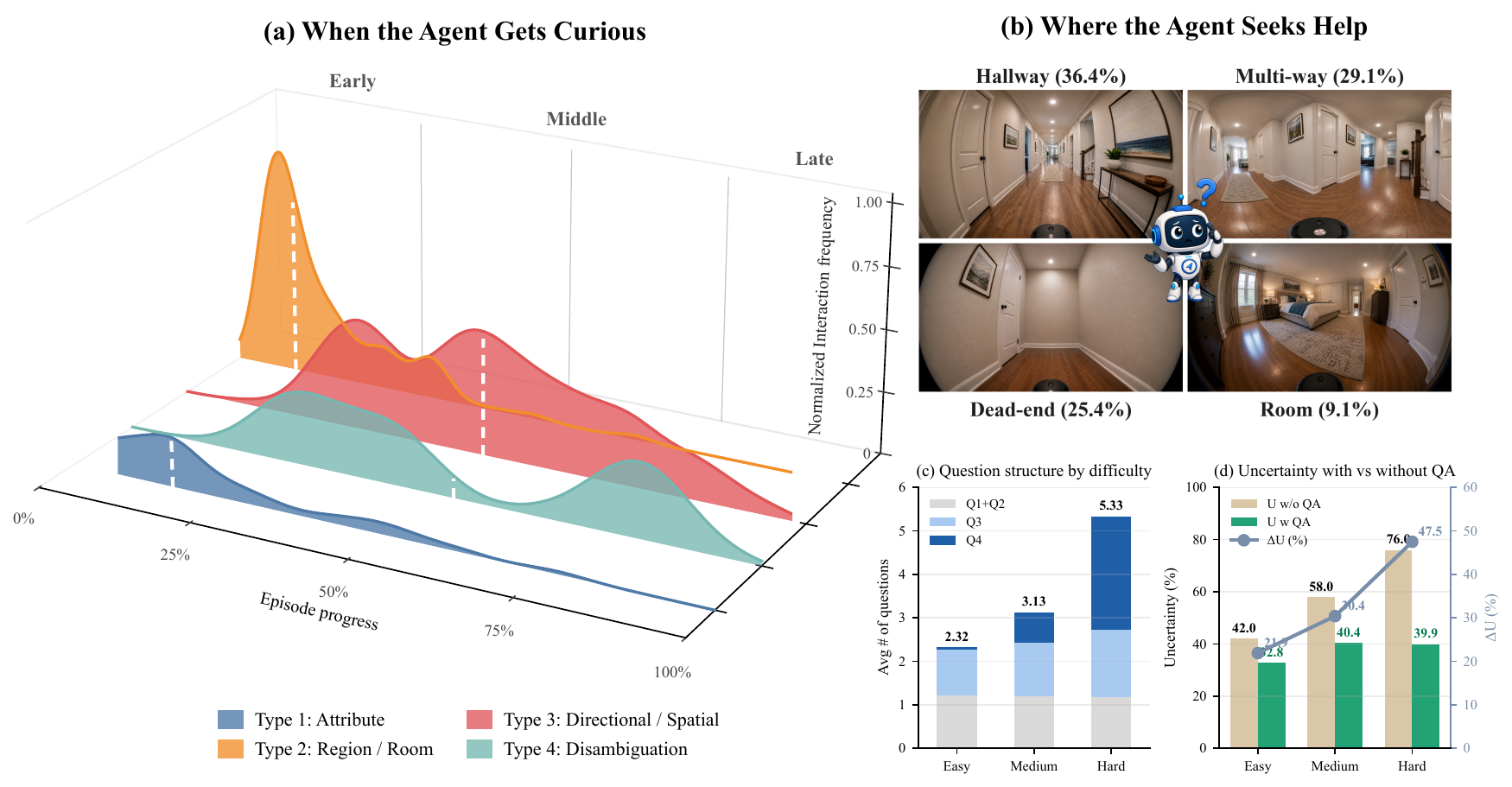}
    \caption{Temporal and spatial patterns of interaction for the full TANDEM agent.}
    \label{fig:exp:dynamics}
\end{figure}


\mypara{When the agent gets curious.}
Figure~\ref{fig:exp:dynamics}(a) illustrates the temporal distribution of question types throughout navigation.
Type~1 (appearance) and Type~2 (region) questions are concentrated in the first $30\%$--$40\%$ of the episode, consistent with TANDEM grounding the target identity and goal region before committing to a navigation path. Type~1 questions appear relatively infrequently (about $20\%$ of episodes), suggesting that the benchmark caption is usually sufficient for target grounding and that additional appearance queries are triggered mainly under perceptual uncertainty. Type~4 (confirmation) exhibits a bimodal pattern, with a smaller mid-episode peak around $30\%$ caused by candidate verification in incorrect rooms, followed by a dominant late-stage peak near $80\%$ when the agent checks whether it has reached the correct target. In contrast, Type~3 (direction/route) questions remain comparatively uniform throughout the episode, suggesting that spatial confusion can arise at any point in the episode.



\mypara{When the agent seeks help.}
Figure~\ref{fig:exp:dynamics}(b) shows where Type~3 questions arise during navigation. These questions concentrate at long corridors, multi-way junctions, and dead-ends, where local first-person observations are insufficient to infer the global layout. In contrast, Type~3 questions occur inside closed rooms only $7\%$ of the time, indicating that TANDEM usually handles local within-room search without spending route-level interaction budget. Among Type~3 questions, only $10\%$ are explicit route requests, while $65\%$ ask about the spatial position of the goal room and the remaining $25\%$ concern relations between rooms. This pattern suggests that Type~3 interaction primarily provides coarse spatial guidance rather than a substitute for navigation itself.


\mypara{Interaction scales with uncertainty.}
Figure~\ref{fig:exp:dynamics}(c) and (d) analyze how interaction varies across difficulty levels and quantify the uncertainty reduction associated with Type~3 questions. The average number of questions $|Q|$ increases from $2.32$ on easy episodes to $5.33$ on hard ones. This increase is driven primarily by Type~3 questions ($1.05 \rightarrow 1.56$) and especially Type~4 questions ($0.05 \rightarrow 2.60$). The sharp rise in Type~4 usage correlates with the number of distractors in harder episodes, highlighting why a uniform-cost interaction metric would disproportionately penalize difficult cases. To quantify the effect of Type~3 interaction, we use explored area as a proxy for navigation uncertainty. Let $A_{\mathrm{noQA}}$ denote the area explored before a Type~3 question and $A_{\mathrm{QA}}$ the area explored over a matched horizon after incorporating the answer. We define uncertainty reduction as $\Delta U = (A_{\mathrm{noQA}} - A_{\mathrm{QA}})/A_{\mathrm{noQA}}$. Using this proxy, $\Delta U$ increases from $21.9\%$ on easy episodes to $47.5\%$ on hard episodes, suggesting that Type~3 interaction becomes increasingly effective at compressing the search space as layout ambiguity grows.



Overall, the results suggest that different question types play complementary roles throughout navigation. Appearance and region questions reduce early-stage target uncertainty, route/layout questions resolve spatial ambiguity when local observations are insufficient, and confirmation questions help avoid stopping at distractors near the end of an episode. Together with the dedicated question policy and cost-aware metric, TANDEM concentrates interaction at high-uncertainty moments, uses only a small number of questions, and achieves the largest uncertainty reduction on hard episodes. These behavioral patterns are consistent with the performance gains observed in the ablations of Table~\ref{tab:agent_results}.

\subsection{Case Study}
\label{sec:exp:case}

\begin{figure}[h]
    \centering
    \includegraphics[width=0.8\linewidth]{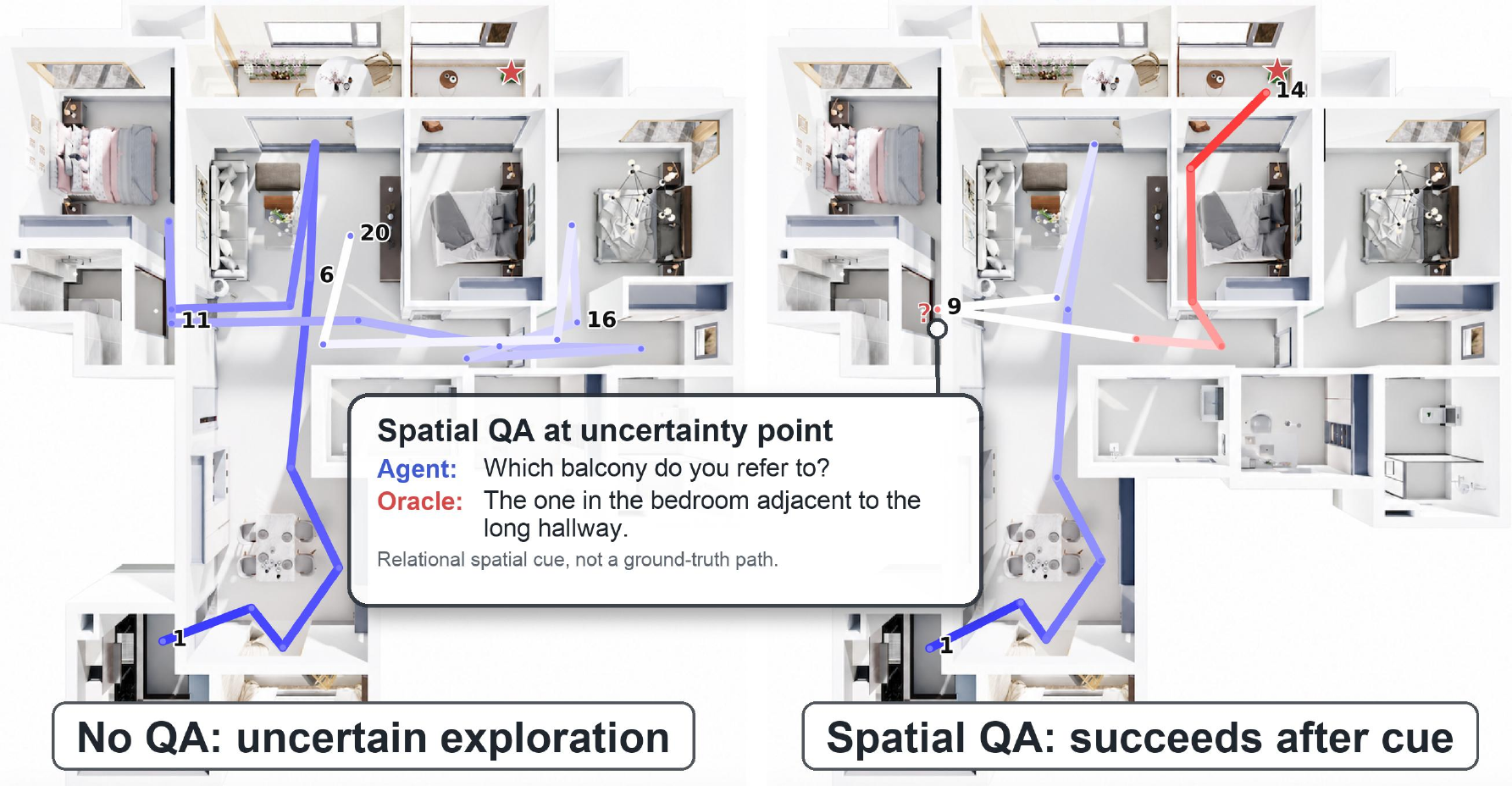}
    \caption{\textbf{Effect of spatial interaction.} A spatial QA cue helps the agent resolve ambiguity and reach the target more directly, instead of following uncertain exploratory paths.}
    \label{fig:exp:case0010}
\end{figure}


Figure~\ref{fig:exp:case0010} presents a paired episode in which the full TANDEM agent succeeds while the no-QA ablation fails under the same start state and target. The no-QA agent exhibits competent local reasoning: it identifies plausible candidate rooms and executes a multi-room search, but converges on the wrong wing of the layout.
A single Type~3 question at the marked junction redirects the agent toward the correct wing, after which it quickly reaches the target room. Importantly, the Oracle response is a natural-language route hint, such as ``head past the dining area, then look for the room with the desk on the left at the end of the hallway,'' rather than explicit geometric supervision.
The gain therefore comes from refining the agent's coarse spatial prior rather than providing direct navigation instructions.
A second staircase-junction example with the same paired structure is provided in the appendix.

%% file: sec/tab/agent_table.tex
\begin{table*}[t!]
\caption{Naive and cost-aware QA agent comparison on the evaluation subset with Qwen3.5-8B.}
\label{tab:agent_results}
\centering
\renewcommand{\arraystretch}{1.05}
\resizebox{\linewidth}{!}{
\begin{tabular}{>{\raggedright\arraybackslash}p{4.4cm}|*{2}{c}|*{2}{c}|*{2}{c}|*{5}{c}|*{7}{c}}
\toprule
\multirow{3}{*}{\textbf{Agent / Variant}}
& \multicolumn{11}{c|}{\cellcolor{orange!10}\textbf{Navigation Performance}}
& \multicolumn{7}{c}{\cellcolor{cyan!10}\textbf{Dialog \& Cost}} \\
\cmidrule(lr){2-12}\cmidrule(lr){13-19}
& \multicolumn{2}{c|}{\textit{Easy}}
& \multicolumn{2}{c|}{\textit{Medium}}
& \multicolumn{2}{c|}{\textit{Hard}}
& \multicolumn{5}{c|}{\textit{Overall}}
& \multicolumn{7}{c}{} \\
\cmidrule(lr){2-3}\cmidrule(lr){4-5}\cmidrule(lr){6-7}\cmidrule(lr){8-12}
& SR$@$1.5 & $|Q|$ & SR$@$1.5 & $|Q|$ & SR$@$1.5 & $|Q|$ & SR/OSR$@$0.5 & SR/OSR$@$1.5 & SR/OSR$@$3 & NE & TL & $Q_1$ & $Q_2$ & $Q_3$ & $Q_4$ & $|Q|$ & WSR$@$1.5 & WSR$@$3 \\
\midrule
\rowcolor{gray!12}\multicolumn{19}{l}{\textbf{Naive QA Baselines (unconstrained interaction)}} \\
\midrule
GTA~\cite{li2026gta} & 29.7 & 6.93 & 27.6 & 8.17 & 22.4 & 9.76 & 8.8/17.6 & 26.7/39.3 & 50.6/57.9 & 3.49 & 27.62 & 1.73 & 1.46 & 2.87 & 2.19 & 8.25 & 6.4 & 12.2 \\
COIN~\cite{taioli2025coin} & 23.6 & 7.48 & 18.7 & 9.03 & 14.3 & 10.81 & 5.9/12.4 & 18.6/29.2 & 36.8/46.3 & 4.63 & 25.18 & 2.04 & 1.17 & 3.58 & 2.27 & 9.06 & 3.7 & 7.3 \\
MapGPT~\cite{chen2024mapgpt} & 20.1 & 7.27 & 17.1 & 8.91 & 12.7 & 10.46 & 6.7/12.1 & 15.9/28.3 & 32.6/41.7 & 4.89 & 29.07 & 1.84 & 1.39 & 3.11 & 2.53 & 8.87 & 3.6 & 7.1 \\
NavGPT~\cite{zhou2024navgpt} & 18.1 & 7.34 & 15.6 & 8.99 & 10.9 & 10.68 & 5.1/11.4 & 14.8/24.6 & 28.9/36.7 & 5.23 & 31.64 & 1.99 & 1.18 & 3.24 & 2.61 & 9.02 & 3.1 & 6.1 \\
\midrule
\rowcolor{gray!12}\multicolumn{19}{l}{\textbf{Cost-Aware QA Protocol (ours)}} \\
\midrule
GTA~\cite{li2026gta} & 29.2 & 2.24 & 27.0 & 3.01 & 21.8 & 5.14 & 10.5/17.2 & 26.0/38.6 & 49.3/56.1 & 3.55 & 26.85 & 0.24 & 0.95 & 1.20 & 1.05 & 3.44 & 15.9 & 30.3 \\
MapGPT~\cite{chen2024mapgpt} & 19.6 & 2.15 & 17.4 & 2.87 & 12.2 & 4.87 & 6.6/12.3 & 16.5/27.6 & 31.5/40.2 & 4.97 & 28.68 & 0.26 & 0.93 & 1.10 & 0.99 & 3.28 & 10.2 & 19.4 \\
NavGPT~\cite{zhou2024navgpt} & 17.5 & 2.11 & 15.3 & 2.82 & 10.1 & 4.76 & 5.8/10.8 & 14.3/24.1 & 27.8/35.3 & 5.31 & 29.15 & 0.27 & 0.92 & 1.06 & 0.96 & 3.21 & 8.9 & 16.9 \\
COIN~\cite{taioli2025coin} & 18.0 & 1.37 & 12.5 & 2.19 & 8.5 & 4.33 & 2.0/3.5 & 13.0/22.0 & 27.4/35.9 & 5.20 & 28.40 & 0.45 & 1.00 & 0.00 & 1.16 & 2.61 & 9.2 & 19.3 \\
\textbf{TANDEM} & \cellcolor{teal!25}\textbf{39.0} & 2.32 & \cellcolor{teal!25}\textbf{35.8} & 3.13 & \cellcolor{teal!25}\textbf{31.0} & 5.33 & \cellcolor{teal!25}\textbf{14.2/22.3} & \cellcolor{teal!25}\textbf{35.3/50.0} & \cellcolor{teal!25}\textbf{66.5/72.5} & \cellcolor{teal!25}\textbf{2.68} & 23.87 & 0.22 & 0.97 & 1.28 & 1.10 & 3.57 & \cellcolor{teal!25}\textbf{21.4} & \cellcolor{teal!25}\textbf{40.8} \\
\quad w/o spmem & \cellcolor{teal!7}\textbf{31.2} & 2.10 & \cellcolor{teal!7}\textbf{29.0} & 2.49 & \cellcolor{teal!7}\textbf{24.0} & 4.56 & \cellcolor{teal!7}\textbf{8.8/16.0} & \cellcolor{teal!7}\textbf{28.0/43.5} & \cellcolor{teal!7}\textbf{57.8/65.0} & \cellcolor{teal!7}\textbf{3.15} & 24.08 & 0.26 & 1.00 & 0.82 & 0.93 & 3.01 & \cellcolor{teal!7}\textbf{18.5} & \cellcolor{teal!7}\textbf{36.0} \\
\quad w/o QA & 24.5 & - & 17.0 & - & 17.8 & - & 3.2/10.5 & 20.0/34.0 & 49.2/58.4 & 3.35 & \cellcolor{teal!25}\textbf{17.27} & - & - & - & - & - & 14.2 & 34.2 \\
\quad w/o spmem \& QA & 18.0 & - & 14.0 & - & 10.0 & - & 1.5/6.0 & 14.0/25.0 & 34.2/44.1 & 5.15 & \cellcolor{teal!7}\textbf{21.16} & - & - & - & - & - & 10.0 & 23.8 \\
\bottomrule
\end{tabular}}
\vspace{2pt}\\

\vspace{-6pt}
\end{table*}

%% file: sec/tab/backbone_table.tex
\begin{table*}[t!]
\caption{Navigator backbone comparison for TANDEM on the evaluation subset.}
\label{tab:backbone_results}
\centering
\renewcommand{\arraystretch}{1.05}
\resizebox{\linewidth}{!}{
\begin{tabular}{>{\raggedright\arraybackslash}p{4.4cm}|*{2}{c}|*{2}{c}|*{2}{c}|*{5}{c}|*{6}{c}}
\toprule
\multirow{3}{*}{\textbf{Navigator Backbone}}
& \multicolumn{11}{c|}{\cellcolor{orange!10}\textbf{Navigation Performance}}
& \multicolumn{6}{c}{\cellcolor{cyan!10}\textbf{Dialog \& Cost}} \\
\cmidrule(lr){2-12}\cmidrule(lr){13-18}
& \multicolumn{2}{c|}{\textit{Easy}}
& \multicolumn{2}{c|}{\textit{Medium}}
& \multicolumn{2}{c|}{\textit{Hard}}
& \multicolumn{5}{c|}{\textit{Overall}}
& \multicolumn{6}{c}{} \\
\cmidrule(lr){2-3}\cmidrule(lr){4-5}\cmidrule(lr){6-7}\cmidrule(lr){8-12}
& SR$@$1.5 & $|Q|$ & SR$@$1.5 & $|Q|$ & SR$@$1.5 & $|Q|$ & SR/OSR$@$0.5 & SR/OSR$@$1.5 & SR/OSR$@$3 & NE & TL & $Q_1$ & $Q_2$ & $Q_3$ & $Q_4$ & WSR$@$1.5 & WSR$@$3 \\
\midrule
\rowcolor{gray!12}\multicolumn{18}{l}{\textbf{Closed-Source MLLMs}} \\
\midrule
GPT-5.4~\cite{openai2026gpt54} & \cellcolor{teal!7}\textbf{45.5} & 2.33 & \cellcolor{teal!25}\textbf{42.2} & 3.22 & \cellcolor{teal!7}\textbf{37.2} & 5.58 & \cellcolor{teal!7}\textbf{19.2/27.4} & \cellcolor{teal!25}\textbf{41.6/56.4} & \cellcolor{teal!25}\textbf{74.8/81.0} & \cellcolor{teal!7}\textbf{2.21} & \cellcolor{teal!25}\textbf{23.21} & 0.20 & 0.98 & 1.33 & 1.18 & \cellcolor{teal!25}\textbf{25.0} & \cellcolor{teal!25}\textbf{45.3} \\
Gemini3-Flash~\cite{google2025gemini3} & \cellcolor{teal!25}\textbf{46.1} & 2.61 & \cellcolor{teal!7}\textbf{41.4} & 3.36 & \cellcolor{teal!25}\textbf{38.2} & 5.69 & \cellcolor{teal!25}\textbf{20.7/28.1} & \cellcolor{teal!7}\textbf{41.1/59.4} & \cellcolor{teal!7}\textbf{73.7/84.2} & \cellcolor{teal!25}\textbf{2.18} &\cellcolor{teal!7}\textbf{19.35} & 0.42 & 0.95 & 1.25 & 1.20 & \cellcolor{teal!7}\textbf{23.8} & \cellcolor{teal!7}\textbf{44.5} \\
\midrule
\rowcolor{gray!12}\multicolumn{18}{l}{\textbf{Open-Source MLLMs}} \\
\midrule
Qwen3.5-8B~\cite{qwen2026qwen35} & 39.0 & 2.32 & 35.8 & 3.13 & 31.0 & 5.33 & 14.2/22.3 & 35.3/50.0 & 66.5/72.5 & 2.68 & 23.87 & 0.22 & 0.97 & 1.28 & 1.10 & 21.4 & 40.7 \\
Qwen3-VL-4B~\cite{bai2025qwen3vl} & 27.4 & 2.26 & 24.6 & 3.05 & 20.4 & 5.22 & 5.4/16.0 & 23.9/39.8 & 51.8/60.6 & 2.97 & 24.32 & 0.19 & 1.00 & 1.23 & 1.07 & 17.7 & 37.4 \\
Qwen3.5-4B~\cite{qwen2026qwen35} & 25.2 & 2.21 & 22.4 & 2.97 & 18.2 & 5.04 & 4.9/14.5 & 22.0/37.9 & 47.2/56.0 & 3.21 & 24.72 & 0.25 & 0.95 & 1.16 & 1.03 & 13.4 & 29.5 \\
Gemma-e4B~\cite{deepmind2026gemma4} & 22.5 & 2.14 & 19.7 & 2.86 & 15.5 & 4.84 & 4.3/13.0 & 19.3/34.2 & 41.4/50.5 & 3.59 & 25.36 & 0.27 & 0.93 & 1.08 & 0.98 & 12.0 & 26.2 \\
InternVL3.5-4B~\cite{wang2025internvl35} & 21.0 & 2.09 & 18.2 & 2.80 & 14.1 & 4.72 & 4.0/12.1 & 17.8/31.8 & 38.2/47.0 & 3.80 & 25.71 & 0.28 & 0.92 & 1.04 & 0.95 & 11.2 & 24.4 \\
\bottomrule
\end{tabular}}
\vspace{2pt}\\
\vspace{-6pt}
\end{table*}

%% file: sec/tab/simulator_result.tex
\begin{wraptable}{r}{0.4\linewidth}
\vspace{-2pt}
\centering
\caption{Agents performance across CVDN, SOON, and REVERIE tasks in discrete environments.}
\label{tab:discrete_filtered}
\vspace{2pt}
\renewcommand{\arraystretch}{1.0}
\setlength{\tabcolsep}{2pt}
\resizebox{\linewidth}{!}{
\definecolor{Gray}{gray}{0.94}
\begin{tabular}{lcccccccccc}
\toprule
\midrule
\multicolumn{1}{c}{\multirow{3}{*}{Methods}} 
& \multicolumn{2}{c}{CVDN} & \multicolumn{4}{c}{SOON} & \multicolumn{4}{c}{REVERIE} \\ 
\cmidrule(r){2-3}
\cmidrule(r){4-7}
\cmidrule(r){8-11}
 & \multicolumn{1}{c}{Val} & \multicolumn{1}{c}{Test} 
 & \multicolumn{2}{c}{Val unseen} & \multicolumn{2}{c}{Test unseen} 
 & \multicolumn{2}{c}{Val unseen} & \multicolumn{2}{c}{Test unseen} \\
 \cmidrule(r){2-2} \cmidrule(r){3-3} 
 \cmidrule(r){4-5} \cmidrule(r){6-7} 
 \cmidrule(r){8-9} \cmidrule(r){10-11}
 & GP $\uparrow$ & GP $\uparrow$ 
 & SR $\uparrow$ & SPL $\uparrow$ & SR $\uparrow$ & SPL $\uparrow$ 
 & SR $\uparrow$ & SPL $\uparrow$ & SR $\uparrow$ & SPL $\uparrow$ \\ 
\midrule
\midrule
HAMT~\cite{chen2021hamt}
& 5.13 & 5.58 & - & - & - & - & 33.0 & 30.2 & 30.4 & 26.7 \\
DUET~\cite{chen2022duet}
& - & - & 36.3 & 22.6 & 33.4 & 21.4 & 47.0 & 33.7 & 52.5 & 36.1 \\
AutoVLN~\cite{chen2022hm3dlearning}
& - & - & 41.0 & 30.7 & 40.4 & 27.9 & 55.9 & 40.9 & 55.2 & 38.9 \\
GOAT~\cite{wang2024vision}
& - & - & 40.4 & 28.1 & 40.5 & 25.2 & 53.4 & 36.7 & 57.7 & 40.5 \\
ScaleVLN~\cite{wang2023scaling}
& 6.12 & 6.97 & - & - & - & - & 57.0 & 41.8 & 56.1 & 39.5 \\
NaviLLM~\cite{zheng2023towards}
& 6.16 & 7.90 & 38.3 & 29.2 & 35.0 & 26.3 & 42.2 & 35.7 & 39.8 & 32.3 \\
SAME~\cite{zhou2025same}
& 6.94 & 7.07 & 36.1 & 25.4 & 38.2 & 27.1 & 46.4 & 36.1 & 48.6 & 37.1 \\
\midrule
TANDEM
& \textbf{8.15} & \textbf{8.50} & \textbf{45.2} & \textbf{33.5} & \textbf{44.8} & \textbf{31.2} & \textbf{62.4} & \textbf{45.6} & \textbf{63.1} & \textbf{44.2} \\
\bottomrule
\end{tabular}}
\label{tab:simulator_compact}
\vspace{-8pt}
\end{wraptable}

%% file: sec/5_conclusion.tex
\section{Conclusion}
\label{sec:conclusion}

We studied interactive Instance Goal Navigation as cost-sensitive uncertainty reduction. Our benchmark makes instance-level ambiguity controllable, grounds oracle answers by question type, and evaluates agents with Weighted Success Rate so that success is judged together with the information consumed. TANDEM provides a strong zero-shot baseline under this protocol by combining cost-aware querying, structured spatial memory, visual grounding, and deterministic metric execution. Experiments show that interaction is most useful when it is typed and selectively used, especially in hard episodes with distractors or layout ambiguity. We also discuss a real-world deployment pipeline, limitations, and broader impact in Appendices~\ref{app:real-world}, \ref{app:limitations}, and~\ref{app:broader-impact}.

%% file: sec/6_appendix.tex
\section*{Appendix}
\label{app:appendix}

\section{Uncertainty Mining and Question Penalties}
\label{app:uncertainty}

\subsection{Annotation Sources and Protocol}
The uncertainty model is built as a reproducible annotation-and-aggregation study. Raw VLN annotations are collected, normalised into shared record formats, source-balanced by downsampling, labelled by Qwen3.5-4B under a fixed schema, and then aggregated into cue-level and type-level statistics. The raw annotations come from R2R~\cite{anderson2018r2r}, REVERIE~\cite{qi2020reverie}, RxR~\cite{anderson2020rxr}, CVDN~\cite{thomason2020cvdn}, and SOON~\cite{zhu2021soon}. Model annotation is kept separate from metric construction: the model never sees our agent trajectories or success labels, and all penalty values can be recomputed from the saved schema outputs without rerunning navigation.

The canonicalisation step uses three input formats because existing VLN data exposes priors in different forms. Pure-text records contain one navigation instruction; dialog records contain a target object and navigator/oracle turns; object records contain structured descriptions with attribute, relation, region, neighbouring-region, and full-description fields. The five source pools are not naturally balanced, so using them directly would let the largest route-instruction corpora dominate the cue statistics. We therefore downsample each dataset to the size of the smallest canonical pool, $3{,}677$ analysis units, with a fixed random seed before annotation. Table~\ref{tab:app:uncertainty-inputs} reports the balanced subset used for uncertainty mining. The counts are at the analysis-unit level: multi-instruction navigation items are expanded into individual instructions, while repeated SOON object descriptions are deduplicated before the balancing step.

\begin{table}[h]
    \centering
    \small
    \caption{Source-balanced annotation subset used for uncertainty mining.}
    \label{tab:app:uncertainty-inputs}
    \begin{tabular}{p{0.50\linewidth}p{0.30\linewidth}r}
        \toprule
        Source family & Canonical unit & Units \\
        \midrule
        R2R route instructions & pure-text instruction & 3,677 \\
        REVERIE object-grounded instructions & pure-text/object instruction & 3,677 \\
        RxR multilingual route instructions & pure-text instruction & 3,677 \\
        CVDN navigation dialogs & dialog history & 3,677 \\
        SOON object descriptions & structured object record & 3,677 \\
        \midrule
        Total & N/A & 18,385 \\
        \bottomrule
    \end{tabular}
\end{table}

For the actual uncertainty run, Qwen3.5-4B is used as the labelling model. The model annotates only this balanced $18{,}385$-unit subset, and all cue rates, count statistics, and locked type-cost aggregation below are computed from the same balanced subset. The downsampling is performed before any agent evaluation and is never recomputed using our evaluation episodes, agent outcomes, or question logs. The model identifies which prior cues are present, how important each cue is, and whether the cue would reduce search uncertainty. These labels are used as observable proxies for useful navigation priors, not as estimates of the true posterior information gain in our benchmark scenes. The labelling input is restricted to textual or structured prior information from existing VLN annotations; it does not include our evaluation episodes, agent actions, oracle answers, success outcomes, or question counts. Thus, the mined uncertainty scores cannot leak test-set navigation performance into the benchmark metrics.

We use Qwen3.5-4B for this step because the task is not to solve navigation but to consistently label semantic priors. A smaller or weaker annotator would make the cue taxonomy noisier, while a closed proprietary annotator would make the penalty construction harder to reproduce. The annotation prompt is fixed across all datasets and uses the same schema for instructions, dialogs, and object descriptions; only the input record changes. Per-record cue annotations are produced first, and the question-type penalties are computed only after aggregation, so the penalty values are not tuned against any agent run.

A natural concern is that the uncertainty annotations may inherit biases from Qwen3.5-4B. We treat this model only as a fixed annotation instrument, not as an evaluator or online decision maker. The annotator never observes our benchmark episodes, agent trajectories, oracle answers, question counts, or success labels, and therefore cannot leak test-set performance into the metric. Its outputs are further constrained by a closed cue schema and evidence requirement, then aggregated over the source-balanced annotation subset before being collapsed into four type-level costs. Model-specific annotation noise can affect the absolute calibration of the costs, but it cannot tune costs to a particular episode or agent variant. All methods are evaluated with the same locked weights, so the penalties serve as reproducible, corpus-derived proxies for relative information value rather than human-ground-truth estimates of information gain.

\subsection{Cue Taxonomy and Aggregation}
The scorer is constrained by a closed output schema rather than asked for free-form rationales. Each record must contain (i) whether disambiguation is needed, (ii) a list of mentioned priors, each with evidence, an importance score in $[0,1]$, and an estimated entropy-reduction level in \{low, medium, high\}, and (iii) a ranked prior list. The 17 cue categories are grouped as in Table~\ref{tab:app:cue-taxonomy}. The taxonomy separates target identity, room context, route structure, spatial relations, visual attributes, and explicit disambiguation. This separation matters because these priors have different leakage profiles: a color cue helps identify an instance, a room cue narrows search space, and a route cue can directly shape navigation. Keeping them separate prevents a single broad ``helpfulness'' label from hiding these differences.

The closed schema makes outputs aggregatable across datasets, prevents Qwen3.5-4B from inventing a different ontology for each corpus, and forces every cue assignment to be tied to evidence in the input record. If a record contains no evidence for a cue, the cue should not be counted. The resulting mining process remains auditable and avoids rewarding plausible but unsupported priors.

\begin{table}[h]
    \centering
    \small
    \caption{Closed prior-cue taxonomy used by the uncertainty scorer.}
    \label{tab:app:cue-taxonomy}
    \begin{tabular}{lp{0.68\linewidth}}
        \toprule
        Group & Cue categories \\
        \midrule
        Target & \texttt{target\_identity}, \texttt{endpoint} \\
        Room & \texttt{room\_name}, \texttt{room\_region}, \texttt{room\_to\_room\_relation} \\
        Navigation & \texttt{landmark}, \texttt{direction\_action}, \texttt{path\_sequence} \\
        Spatial & \texttt{spatial\_relation}, \texttt{support\_placement}, \texttt{object\_location} \\
        Attribute & \texttt{color}, \texttt{shape}, \texttt{size}, \texttt{material}, \texttt{state} \\
        Meta & \texttt{disambiguation\_cue} \\
        \bottomrule
    \end{tabular}
\end{table}

After annotation, each cue $c$ is summarized by four quantities. All reported count statistics use the balanced annotation subset, with $N=18{,}385$ records and $135{,}128$ cue mentions:
\begin{align}
    m_c &= |\{j : \mathrm{cue}(j)=c\}|, &
    \bar{s}_c &= \frac{1}{m_c}\sum_{j:\mathrm{cue}(j)=c} s_j, \nonumber\\
    h_c &= \frac{|\{j : \mathrm{cue}(j)=c,\ \mathrm{gain}(j)=\mathrm{high}\}|}{m_c}, &
    r_c &= \frac{|\{i : \mathrm{rank1}(i)=c\}|}{N},
    \label{eq:app:cue-stats}
\end{align}
where $s_j$ is the Qwen3.5-4B-assigned importance score, $h_c$ is the high-gain ratio among mentions, $r_c$ is the fraction of records in which $c$ is ranked first, and $N=18{,}385$ is the balanced subset size. Disambiguation is marked as needed in $12{,}135$ records ($66.0\%$), confirming that the mined corpora contain substantial ambiguity rather than only fully specified route following.

We use multiple statistics rather than raw frequency alone because frequency is an unsafe proxy for utility. Common cues such as landmarks may appear often but be weakly disambiguating, while target identity or endpoint cues may appear less broadly yet carry high uncertainty reduction. Combining mention count, average importance, high-gain ratio, and top-1 frequency reduces both failure modes: over-penalising common low-value cues and under-penalising rare but highly revealing cues.

\begin{table*}[t]
    \centering
    \small
    \caption{Most frequent cue categories in the source-balanced uncertainty-mining output.}
    \label{tab:app:cue-leaderboard}
    \resizebox{\linewidth}{!}{
    \begin{tabular}{lrrrr|lrrrr}
        \toprule
        Cue & Mentions & Avg. score & High gain & Top-1 &
        Cue & Mentions & Avg. score & High gain & Top-1 \\
        \midrule
        \texttt{direction\_action} & 16,957 & 0.7975 & 59.6\% & 5,567 &
        \texttt{landmark} & 9,476 & 0.6367 & 5.5\% & 67 \\
        \texttt{object\_location} & 12,784 & 0.6578 & 21.6\% & 130 &
        \texttt{target\_identity} & 9,125 & 0.9067 & 90.4\% & 7,803 \\
        \texttt{room\_region} & 11,183 & 0.6581 & 7.5\% & 125 &
        \texttt{path\_sequence} & 7,813 & 0.7905 & 66.4\% & 952 \\
        \texttt{endpoint} & 10,870 & 0.8444 & 78.4\% & 2,760 &
        \texttt{room\_to\_room\_relation} & 7,649 & 0.6661 & 10.0\% & 154 \\
        \texttt{room\_name} & 10,245 & 0.7319 & 22.3\% & 385 &
        \texttt{color} & 7,005 & 0.6212 & 12.9\% & 255 \\
        \texttt{spatial\_relation} & 9,976 & 0.5766 & 2.4\% & 5 &
        \texttt{disambiguation\_cue} & 6,269 & 0.5985 & 22.9\% & 154 \\
        \bottomrule
    \end{tabular}}
\end{table*}

The 17 mined cues are then mapped to the four question types used by the agent and metric. Type~1 asks for appearance and instance identity cues; Type~2 asks where the target is; Type~3 asks for route or coarse-layout information; Type~4 asks whether a currently observed candidate is the target. The collapse in Table~\ref{tab:app:cue-collapse} is fixed before evaluation and is based on the semantics of the answer, not on the performance of any agent variant. Since changing this mapping after seeing results would make Weighted SR tunable, we define the mapping from the prior taxonomy to the interaction interface once and apply the same costs to all methods.

\begin{table}[h]
    \centering
    \small
    \caption{Mapping from mined cue categories to interaction-level question types.}
    \label{tab:app:cue-collapse}
    \begin{tabular}{lp{0.62\linewidth}}
        \toprule
        Question type & Cue categories \\
        \midrule
        Type~1: appearance & \texttt{target\_identity}, \texttt{color}, \texttt{shape}, \texttt{size}, \texttt{material}, \texttt{state} \\
        Type~2: location/region & \texttt{room\_name}, \texttt{room\_region}, \texttt{object\_location}, \texttt{spatial\_relation}, \texttt{support\_placement}, \texttt{endpoint} \\
        Type~3: direction/route & \texttt{direction\_action}, \texttt{path\_sequence}, \texttt{room\_to\_room\_relation}, \texttt{landmark} \\
        Type~4: disambiguation & \texttt{disambiguation\_cue} \\
        \bottomrule
    \end{tabular}
\end{table}

For the final question penalties, each cue is first assigned a composite utility score
\begin{equation}
    U_c = \lambda_1 \bar{s}_c + \lambda_2 h_c + \lambda_3 r_c ,
    \label{eq:app:cue-utility}
\end{equation}
using the three observable proxies in Eq.~\eqref{eq:app:cue-stats}: average scorer confidence, high-gain frequency, and rank-1 decisiveness. We then aggregate from cues to a question type by a mention-weighted average,
\begin{equation}
    U_t = \frac{\sum_{c\in T_t} m_c U_c}{\sum_{c\in T_t} m_c}.
    \label{eq:app:type-utility}
\end{equation}
Finally, the four $U_t$ values are linearly rescaled into the penalty range used by the experiments, with larger utilities mapped to larger costs because they correspond to stronger oracle assistance. The locked penalties are
\[
    (w_1,w_2,w_3,w_4) = (0.182,\ 0.162,\ 0.240,\ 0.103),
\]
for Type~1/2/3/4 respectively. Type~3 is deliberately most expensive because route and layout answers have the highest leakage risk; Type~4 is cheapest because it verifies a local candidate and does not by itself reveal a route.

The penalty design is intentionally monotonic and type-level rather than episode-specific. We do not assign lower costs to questions that happen to help our agent on a given episode, nor do we let the controller choose its own penalty. Every repeated question is charged again, and every method is scored by the same locked weights. This makes the metric conservative: interaction is rewarded only when it improves success enough to compensate for the information it consumes.

Given these locked weights, the reported dataset-level score at threshold $\tau$ is recomputed offline from the stored question-type counts:
\begin{equation}
    \mathrm{WSR}(\tau)
    =
    \frac{1}{|\mathcal{E}|}
    \sum_{e\in\mathcal{E}}
    \mathbf{1}\{d_e \le \tau\}
    \exp\left(-\sum_{t=1}^{4} w_t n_{e,t}\right),
    \label{eq:app:wsr}
\end{equation}
where $\mathcal{E}$ is the evaluation set, $d_e$ is the final distance to the goal, and $n_{e,t}$ is the number of Type-$t$ questions asked in episode $e$. Failed episodes therefore receive zero regardless of question count; repeated questions are charged repeatedly; and the score reduces to standard SR when no questions are asked. All reported values are recomputed after evaluation from stored per-episode question-type counts, rather than taken from the online controller.

\section{Benchmark Construction Details}
\label{app:benchmark}

\subsection{Episode Definition and Filtering}
The benchmark is an interactive instance-level ObjectNav task. The agent receives a coarse instruction such as ``Find a laptop'' and must use interaction to identify the correct object instance among same-category distractors. Unlike traditional VLN, no ground-truth trajectory annotation is required: each episode is defined by a start pose, a target object instance, a goal pose/radius, reference imagery, an oracle description, and metadata needed for evaluation.

Each final episode contains the scene identifier, difficulty label, coarse instruction, target category, target instance identifier, start pose, goal region, goal room, reference-view identifier, natural-language target description, and a deliberately imperfect oracle-memory hint. Intermediate construction records additionally keep distractor lists, target bounding boxes, and structured description attributes; these internal fields are used for validation and analysis but are removed from the released evaluation subset when they would give the agent unfair access to the answer.

We start from the scene's instance graph and room-object graph. Candidate targets are restricted to a manually checked whitelist of 83 object categories that are discrete, visually identifiable, and natural as ``Find a ...'' goals. Structural or ambiguous categories such as walls, floors, doors, windows, ceiling lights, and generic ``unknown'' objects are excluded. Extremely frequent categories such as books, ornaments, pillows, wine sets, flowers, and plates are capped per scene using a room round-robin rule so that the benchmark does not collapse into a few high-count categories. Before reference-image and description filtering, this selection yields $37{,}106$ candidate targets from 262 scenes, with an average of 142 candidates per scene.

\subsection{Difficulty and Episode Assembly}
The construction-time difficulty score is deterministic and combines four factors:
\begin{equation}
    D =
    \alpha_1 F_{\mathrm{distractor}}
    + \alpha_2 F_{\mathrm{room}}
    + \alpha_3 F_{\mathrm{context}}
    + \alpha_4 F_{\mathrm{start}} ,
    \label{eq:app:difficulty}
\end{equation}
$F_{\mathrm{distractor}}$ increases with the number of same-category instances in the scene, using a log scale capped at 1. $F_{\mathrm{room}}$ increases when same-category distractors lie in the same room as the target, since a room answer alone then cannot disambiguate the object. $F_{\mathrm{context}}=1-\rho$ where $\rho$ is the fraction of the target's non-structural nearby anchors that are unique relative to its distractors; if every nearby anchor is shared, spatial relations are less informative and the episode is harder. $F_{\mathrm{start}}$ is the normalised initial path distance from the start pose to the target, which captures the search burden before any question is asked. All four factors are normalised to $[0,1]$ before combination. The fixed coefficients are $(\alpha_1,\alpha_2,\alpha_3,\alpha_4)=(0.35,0.25,0.15,0.25)$ for distractor count, same-room distractors, contextual ambiguity, and initial path distance respectively. They are fixed before episode sampling, are not tuned using agent trajectories, question counts, or success labels, and sum to one. Scores are binned as easy for $D<0.30$, medium for $0.30\le D<0.60$, and hard for $D\ge0.60$.

Start positions are sampled from navigable points and checked for reachability. Easy episodes may start in the same room if the start is at least $3$\,m from the target; medium and hard episodes prefer cross-room starts to force room-level reasoning. The oracle-memory hint is intentionally noisy: if distractors exist, it uses a same-room distractor when possible and otherwise the nearest distractor; if there is no distractor, it samples a nearby offset around the target. The resulting episodes test robustness to imperfect memory rather than assuming direct access to the target pose.

Each target is paired with a target-facing reference view. A fixed offline Qwen3.5-4B annotator is used to check target visibility and generate one compact target description with visible attributes and spatial relations to nearby objects, while avoiding room names in the image-only description. The annotator output is not a sufficient acceptance condition: the final assembly pass attaches only accepted descriptions to episodes and removes cases with an invalid category, room leakage, an empty description, or a description that does not separate the target from same-category distractors under the stored attributes and relations.

The final eligible corpus contains $22{,}905$ episodes from 262 scenes after reference and description filtering. It covers 70 target categories and 11 normalised goal-room labels; the largest target categories are cabinet (2,550), chair (1,909), painting (1,459), curtain (1,398), and pillow (962). The natural difficulty distribution is easy 6,855 / medium 6,101 / hard 9,949, with an average difficulty score of 0.512 and an average of 10.13 distractors per episode. The main experiments use a fixed 500-episode evaluation subset selected before model evaluation. This subset removes internal distractor and bounding-box fields, balances difficulty approximately as easy:medium:hard = 30:40:30, and keeps scenes, goal rooms, and target categories diverse.

\subsection{Benchmark Geometry and Validation}
The geometry in this subsection is used for benchmark construction and validation, including reference-view filtering and target-mask checks. It is not part of the agent's Planner or Grounder input at evaluation time. The construction pipeline checks that episode ids are unique, reference images pass the visibility filter, descriptions are non-empty, start and goal positions are within navigable bounds, and a feasible path exists from start to goal. The final evaluation still reports multiple success thresholds because the stored goal radius is intentionally generous for oracle and annotation logic, while the paper's navigation metrics need to distinguish loose room-level arrival from instance-level localisation.

For target localisation and reference-view filtering, image observations are converted to world-frame points using the standard pinhole projection model. Let $K$ be the camera intrinsic matrix, $(u,v)$ a pixel coordinate, and $z=D(u,v)$ the depth value at that pixel. The corresponding camera-frame point is
\begin{equation}
    \mathbf{x}_c(u,v)
    =
    z K^{-1}
    \begin{bmatrix}
        u \\ v \\ 1
    \end{bmatrix}.
    \label{eq:app:camera-backproject}
\end{equation}
If the camera-to-world extrinsic is $T_{wc}=[R_{wc}\ \mathbf{t}_{wc};\ \mathbf{0}^{\top}\ 1]$, the world-frame point is
\begin{equation}
    \begin{bmatrix}
        \mathbf{x}_w(u,v) \\ 1
    \end{bmatrix}
    =
    T_{wc}
    \begin{bmatrix}
        \mathbf{x}_c(u,v) \\ 1
    \end{bmatrix},
    \quad
    \mathbf{x}_w(u,v)=R_{wc}\mathbf{x}_c(u,v)+\mathbf{t}_{wc}.
    \label{eq:app:camera-to-world}
\end{equation}
When an extrinsic is represented as a world-to-camera transform $T_{cw}=[R_{cw}\ \mathbf{t}_{cw};\ \mathbf{0}^{\top}\ 1]$, we first invert it:
\begin{equation}
    T_{wc}=T_{cw}^{-1}
    =
    \begin{bmatrix}
        R_{cw}^{\top} & -R_{cw}^{\top}\mathbf{t}_{cw} \\
        \mathbf{0}^{\top} & 1
    \end{bmatrix}.
    \label{eq:app:extrinsic-inverse}
\end{equation}
For a visible target mask $\mathcal{M}$, the target centre used by the construction checks is the robust average of valid back-projected points,
\begin{equation}
    \bar{\mathbf{x}}_w
    =
    \frac{1}{|\mathcal{M}_{+}|}
    \sum_{(u,v)\in\mathcal{M}_{+}}
    \mathbf{x}_w(u,v),
    \quad
    \mathcal{M}_{+}=\{(u,v)\in\mathcal{M}:D(u,v)>0\}.
    \label{eq:app:mask-center}
\end{equation}
These equations are used only to express geometry in a common coordinate frame; the agent itself receives image observations and language feedback rather than privileged target coordinates.

\section{Agent Implementation Details}
\label{app:agent}

\subsection{Planner-Grounder Loop}
The agent used in the main experiments is a two-stage zero-shot MLLM policy with an explicit interaction state. At each decision step, the Planner receives the current observation, the coarse instruction, a compact Spatial Memory, a Fact Base of oracle-confirmed facts, recent movement history, and the remaining question budget. It then emits one structured decision: ask a question, stop, or choose one of the eight egocentric directions with a natural-language subgoal. If it chooses to move, the Grounder receives only the selected view and the subgoal, selects an $8\times8$ grid cell, and the deterministic executor converts that cell into a short local waypoint. After execution, the Planner's belief update is merged back into Spatial Memory. This loop separates semantic planning from visual grounding: the Planner never emits pixels or metric waypoints, and the Grounder never decides the navigation strategy.

\begin{table}[h]
    \centering
    \small
    \caption{Agent state and decision flow. The table describes what the controller maintains and how it is used.}
    \label{tab:app:agent-flow}
    \begin{tabular}{p{0.18\linewidth}p{0.35\linewidth}p{0.36\linewidth}}
        \toprule
        Component & Input state & Output / update \\
        \midrule
        Observation composer & Current RGB views and heading & One labelled 8-direction panel with local grids \\
        Planner & Instruction, 8-direction panel, Spatial Memory, Fact Base, movement history, question budget & Structured decision: direction, subgoal, purpose, question fields, stop flag, belief update \\
        Oracle interface & Question type, current observation, target metadata, current pose & Natural-language answer; duplicate Type-1/2 questions reuse the latest confirmed answer \\
        Fact Base & Oracle answers and question history & Accumulated appearance, location, spatial-hint, and candidate-confirmation facts \\
        Grounder & Selected single view and Planner subgoal & Grid cell $(c,r)$ in the selected view, with a short rationale \\
        Executor & Grounded cell and current pose & Local waypoint, collision-aware move, trajectory update \\
        \bottomrule
    \end{tabular}
\end{table}

The observation design was changed from a four-view panorama to an eight-direction panel because the agent repeatedly needed to decide between diagonal exits, oblique doorways, and side-front corridor branches. Four 90-degree views preserve full coverage but force diagonal openings to sit near image boundaries, where visual grounding is less stable and Planner language such as ``front-left doorway'' has no direct visual slot. Twelve views reduce angular aliasing further, but each view becomes narrower and the composite observation becomes harder for the Planner to scan; it also increases prompt-image clutter without changing the action interface. The eight-direction layout is therefore the best compromise: it gives every common branch direction an explicit cell, keeps each view wide enough for object and doorway context, and lets the Grounder operate on a single $90^\circ$ view after the Planner has chosen a direction.

\begin{table}[h]
    \centering
    \small
    \caption{Observation-granularity ablation under the same backbone and interaction protocol. The 8-direction row is the main setting; the other two rows are reported as controlled observation variants.}
    \label{tab:app:obs-ablation}
    \begin{tabular}{lrrrrrr}
        \toprule
        Observation & SR@1.5 & SR@3 & OSR@1.5 & NE & $|Q|$ & WSR@1.5 \\
        \midrule
        4 views ($90^\circ$ only) & 32.1 & 62.8 & 46.9 & 2.94 & 3.64 & 19.2 \\
        8 directions (ours) & \textbf{35.3} & \textbf{66.5} & \textbf{50.0} & \textbf{2.68} & 3.57 & \textbf{21.4} \\
        12 narrow views & 33.8 & 64.9 & 48.7 & 2.81 & 3.69 & 20.3 \\
        \bottomrule
    \end{tabular}
\end{table}

The final controller uses four practical safeguards that were added after early pilot runs. First, Type-1 and Type-2 questions are encouraged early only when the Fact Base lacks appearance or room information; once answered, they are cached to avoid repeated global questions. Second, Type-3 is treated as a spatial-uncertainty tool: it is preferred at ambiguous branches, repeated loops, long corridors, and cases where the target room is known but the entry direction is unclear. Third, Type-4 confirmation creates a follow-up state: an affirmative answer either stops immediately or requests exactly one closer move before stopping, preventing the agent from continuing to wander after confirming the target. Fourth, a light recovery rule triggers a Type-3 query when the agent has moved for several steps, remains far from the goal, and has not yet asked for spatial guidance. These details are controller-side guardrails; the metric still charges every resulting question with the same locked per-type cost.

\subsection{Spatial Memory and Prompt Templates}
Spatial Memory is a compact language-level map carried between steps. It is not a metric occupancy map; it is the state summary that the Planner sees in the next prompt. The memory has a fixed shape:
\begin{tcolorbox}[title=Spatial Memory state,colback=gray!4,colframe=black!45,boxrule=0.4pt,arc=1mm,left=4pt,right=4pt,top=4pt,bottom=4pt]
\small
\textbf{Current place:} current-room hypothesis; visible room type; heading-level context. \\
\textbf{Target facts:} appearance, category, distinctive attributes, support surface, and room/location facts confirmed by the Oracle. \\
\textbf{Visited areas:} short list of explored rooms or corridor segments, each with salient objects and whether the target was observed. \\
\textbf{Unexplored openings:} visible doors, corridors, stairs, or branch directions that remain plausible. \\
\textbf{Route hints:} active Type~3 guidance, expected room transition, and coarse direction of travel. \\
\textbf{Blocked or failed moves:} recent directions that led to collision, dead ends, loops, or no semantic progress. \\
\textbf{Candidate status:} visible candidate objects, Type~4 confirmations or rejections, and whether the agent should stop after one closer move.
\end{tcolorbox}
The Planner must return a belief update every step, and the controller merges that update into the next Spatial Memory unless the spatial-memory ablation is active. A typical memory after an early Type~1/2 exchange and two moves looks as follows:
\begin{tcolorbox}[title=Example Spatial Memory,colback=gray!4,colframe=black!45,boxrule=0.4pt,arc=1mm,left=4pt,right=4pt,top=4pt,bottom=4pt]
\small
\textbf{Current place:} likely hallway outside a bedroom-like room; front-left shows an open doorway; rear returns to the entrance corridor. \\
\textbf{Target facts:} target is a black office chair; likely in a study or work area; not on a dining table. \\
\textbf{Visited areas:} entrance corridor with cabinet and painting; side hallway with no chair visible. \\
\textbf{Unexplored openings:} front-left doorway with desk-like furniture; right corridor with shelves; rear-left staircase not yet explored. \\
\textbf{Route hints:} Oracle said the target is near a desk in a room off the hallway; current active hint biases movement toward the left-side room. \\
\textbf{Blocked or failed moves:} front was blocked by a table edge; rear-right caused a loop back to the entrance. \\
\textbf{Candidate status:} no confirmed candidate; one dark chair-like object in front-left should trigger Type~4 if visible after entering.
\end{tcolorbox}
The fixed fields make the state auditable: if the agent repeats a question, ignores a route hint, or walks back into a blocked branch, the error can be traced to a specific memory field rather than hidden model state.

The runtime prompt is assembled from five blocks. Blocks 1 to 4 are fixed system instructions that define the Planner's role, observation format, memory, interaction policy, recovery rules, and output contract. Block 5 is the per-step runtime wrapper that fills in the task instruction, Fact Base, active hints, Spatial Memory, movement history, image observation, heading, step count, and remaining question budget. This separation keeps stable policy constraints distinct from episode state while producing one final prompt for each Planner call. The same prompt blocks are used for all reported episodes, agent variants, and backbones. They are not rewritten for individual scenes, failure cases, difficulty bins, or model families after inspecting outcomes.

\begin{tcolorbox}[title={Planner prompt block 1: role, environment, and observation},colback=gray!4,colframe=black!45,boxrule=0.4pt,arc=1mm,left=4pt,right=4pt,top=4pt,bottom=4pt]
\small
You are an expert Vision-Language Navigation (VLN) agent operating in an \textbf{indoor environment}. Your mission is to find a target object based on a vague description by navigating, observing, and strategically asking questions to an Oracle.

\textbf{Your role.} You are a \textbf{high-level reasoning and planning agent}. You do not output low-level motor commands or pixel coordinates. Instead, you output structured semantic decisions: what to do and why. A separate grounding module translates your plan into actual movement.

\textbf{Environment constraints.}
\begin{itemize}[leftmargin=*,nosep]
    \item This is strictly an indoor navigation task between rooms.
    \item \textbf{Avoid going outside.} If you see sky or outdoor scenery, turn around immediately.
    \item Obstacle avoidance is handled by the system: you focus on what to do and where to go.
\end{itemize}

\textbf{Observation: 8-direction view.} You receive a single composite image showing eight camera views arranged around your current position:
\begin{center}
\scriptsize
\begin{tabular}{ccc}
Front-Left ($45^\circ$) & Front ($0^\circ$) & Front-Right ($315^\circ$) \\
Left ($90^\circ$) & You are here & Right ($270^\circ$) \\
Rear-Left ($135^\circ$) & Rear ($180^\circ$) & Rear-Right ($225^\circ$)
\end{tabular}
\end{center}
The center cell shows your orientation with front/back/left/right arrows. Each surrounding cell shows a $90^\circ$ camera view in that direction. Think of this as a top-down map: front is up, left is left, and so on. The current heading is provided in degrees, with $0^\circ$ as north, $90^\circ$ as east, $180^\circ$ as south, and $270^\circ$ as west. Use this to maintain absolute spatial orientation across steps. Movement-history turn signs follow the Isaac convention: positive degree means turn left and negative degree means turn right.
\end{tcolorbox}

\begin{tcolorbox}[title={Planner prompt block 2: memory and interaction},colback=gray!4,colframe=black!45,boxrule=0.4pt,arc=1mm,left=4pt,right=4pt,top=4pt,bottom=4pt]
\small
\textbf{Spatial working memory.} You maintain a spatial working memory across steps. This memory is provided back to you as \emph{Spatial Memory} in each call. It tracks:
\begin{itemize}[leftmargin=*,nosep]
    \item \texttt{target\_description}: what you know about the target's appearance.
    \item \texttt{visited\_areas}: areas explored and their contents.
    \item \texttt{unexplored\_areas}: visible areas you have not visited.
    \item \texttt{current\_room}: your best guess of what room or area you are in.
    \item \texttt{key\_observations}: important things you have noticed.
\end{itemize}
Update this memory in your \texttt{belief\_update} field. Be specific and thorough: this is your only memory across steps.

\textbf{Interactive capabilities.} The instruction is usually vague, for example, ``find a camera''. Use questions only when the answer changes your next decision.

\textbf{Question types you can ask.}
\begin{itemize}[leftmargin=*,nosep]
    \item Type~1, appearance: high-priority early question for target visual cues.
    \item Type~2, location: high-priority early question for target room or area.
    \item Type~3, spatial: ask during navigation when branch or direction choice is ambiguous.
    \item Type~4, candidate check: ask immediately when a visible object might be the target.
\end{itemize}
\end{tcolorbox}

\begin{tcolorbox}[title={Planner prompt block 3: question, hint, and recovery policies},colback=gray!4,colframe=black!45,boxrule=0.4pt,arc=1mm,left=4pt,right=4pt,top=4pt,bottom=4pt]
\small
\textbf{Early-question policy.}
\begin{itemize}[leftmargin=*,nosep]
    \item If Fact Base does not already contain reliable Type~1 or Type~2 information, proactively ask them early.
    \item Default target behavior: ask at least one of Type~1/2 by step 2, and both by step 5 when questions remain.
    \item Prefer asking early over blind exploration when target appearance or location is still uncertain.
    \item Do not ask duplicates if the answer is already confirmed in Fact Base.
    \item After you have reliable Type~1 and Type~2 facts, do not keep asking Type~1/2 unless strong contradiction appears.
    \item Repeated Type~1/2 is low value; spend remaining question budget on Type~3/4.
\end{itemize}

\textbf{Type~3 guidance.}
\begin{itemize}[leftmargin=*,nosep]
    \item Ask Type~3 at junctions or corridor forks when multiple directions are similarly plausible.
    \item Ask Type~3 when loop risk appears, such as revisiting the same junction or direction flips without progress.
    \item If the goal room is known, for example a study, but the entry direction is unclear, this is a strong Type~3 trigger.
    \item Once Type~1/2 are known, Type~3 is usually the highest-value question during navigation.
    \item If you are unsure between moving and asking at an ambiguous branch, prefer asking Type~3.
    \item Type~3 should request coarse spatial constraints or cues, not route steps.
    \item If one option is clearly best from observation and Fact Base, move without asking.
    \item Never ask a question already answered in Fact Base.
\end{itemize}

\textbf{Type~3 spatial-hint mode.} If \emph{Active Spatial Hint} is present, use it as your main direction bias. The controller reports qualitative progress from the current pose to the active hint after each step. Treat the hint as persistent guidance until a newer Type~3 hint replaces it. Use the hint to bias both \texttt{direction} and \texttt{subgoal}, then re-evaluate after each move. This is guidance for planning bias, not a direct straight-line, fixed-distance, or path-following command. Choose short, locally reachable subgoals such as a doorway, corner, or opening that progress toward the hinted side. While the hint is active and the controller reports that the target is still outside the stop eligibility range, your primary objective is to make visible local progress toward the hinted area. If progress gets worse, immediately recover or replan toward the hinted side, or ask Type~3 again only if branch ambiguity remains high. If strong visual evidence contradicts the hint, trust observation.

\textbf{Stop policy.} Final success metrics are computed offline at the reported thresholds. During online navigation, stop only when the target is visually confirmed, a positive Type~4 follow-up says the agent is close enough, or \emph{Active Spatial Hint} reports stop eligible proximity without visual contradiction. Do not over-explore after stop eligibility; prioritize final confirmation and stop.

\textbf{Anti-oscillation using movement history.} Read \texttt{movement\_history} every step to avoid back-and-forth behavior. Repeated front-to-rear or side-to-opposite-side switching indicates oscillation. \texttt{rear}, \texttt{rear-left}, and \texttt{rear-right} are mainly recovery moves, not default exploration moves. Do not choose rear-family directions repeatedly unless there is clear evidence such as a dead end, wrong room, outdoor risk, or hint behind. After one rear-family recovery move, prefer a new front or side branch to make forward progress. If uncertain, ask Type~3 instead of taking another rear-family move.

\textbf{Type~4 follow-up mode.} If the Type~4 follow-up state says ``yes, move closer then stop'', the target is confirmed: do not ask more questions, take exactly one more navigation move to get closer, then stop on the next decision. If it says ``yes'' with no extra instruction, the target is confirmed and close enough, so stop now. During this mode, use task phase \texttt{verify} or \texttt{approach}.
\end{tcolorbox}

\begin{tcolorbox}[title={Planner prompt block 4: output contract},colback=gray!4,colframe=black!45,boxrule=0.4pt,arc=1mm,left=4pt,right=4pt,top=4pt,bottom=4pt]
\small
\textbf{Output format.} The Planner must output strict JSON with no markdown fences.

\textbf{When navigating:}
\scriptsize
\begin{verbatim}
{
  "reasoning": "Step-by-step spatial reasoning about what you see in the
                8 views and where to go",
  "task_phase": "explore | verify | approach | recover",
  "belief_update": {
    "current_room": "...", "target_description": "...",
    "visited_areas": ["..."], "unexplored_areas": ["..."],
    "key_observations": ["..."]
  },
  "direction": "front-left",
  "subgoal": "Target place: the center of the bedroom doorway in Front-Left,
              about 2-3 meters away",
  "purpose": "Oracle said target is in bedroom, Front-Left shows a bedroom door",
  "ask_question": false, "question_type": null, "question": null,
  "stop": false
}
\end{verbatim}
\small

\textbf{When asking an early clarification question, Type~1 or Type~2, with no movement this step:}
\scriptsize
\begin{verbatim}
{
  "reasoning": "I still lack reliable target appearance/location information,
                and asking now will reduce search uncertainty before more
                movement.",
  "task_phase": "explore",
  "belief_update": {
    "current_room": "...", "target_description": "unknown",
    "visited_areas": [], "unexplored_areas": [], "key_observations": []
  },
  "direction": null, "subgoal": null, "purpose": null,
  "ask_question": true, "question_type": 2,
  "question": "Which room or area is the target most likely in?",
  "stop": false
}
\end{verbatim}
\small

\textbf{When stopping:}
\scriptsize
\begin{verbatim}
{
  "reasoning": "The target is close enough under the stop policy and evidence
                confirms this is the target. Stopping now.",
  "task_phase": "approach",
  "belief_update": { ... },
  "direction": null, "subgoal": null, "purpose": null,
  "ask_question": false, "question_type": null, "question": null,
  "stop": true
}
\end{verbatim}
\small

\textbf{Direction and subgoal.} When navigating, the Planner must output both \texttt{direction} and \texttt{subgoal}. The direction must be exactly one of \texttt{front}, \texttt{front-left}, \texttt{left}, \texttt{rear-left}, \texttt{rear}, \texttt{rear-right}, \texttt{right}, and \texttt{front-right}. Prefer non-rear directions for normal exploration; use rear-family directions mainly for recovery or when hint or evidence points behind. The subgoal must describe a place to move to in that view, not just an action verb. Do not output long fixed-distance intents such as ``go 8-10m toward hint direction''. Prefer short local advances to feasible anchors, then update the plan next step.

A separate grounding module sees only that one view and grounds the place description. Good examples use direction \texttt{front-left} with the subgoal ``Target place: center of the bedroom doorway'', or direction \texttt{rear} with the subgoal ``Target place: hallway opening behind me''. Bad examples include ``turn left and move'', which is action-only, and ``explore'', which is too vague and lacks a direction.

\textbf{Critical rules.} Ask proactively, especially early. Maintain Spatial Memory. Prevent oscillation using movement history. Confirm before stopping. Stay indoors. Use heading to maintain spatial awareness and give directionally accurate subgoal instructions.
\end{tcolorbox}

\begin{tcolorbox}[title={Planner prompt block 5: runtime wrapper},colback=gray!4,colframe=black!45,boxrule=0.4pt,arc=1mm,left=4pt,right=4pt,top=4pt,bottom=4pt]
\small
At each step, the fixed system prompt above is followed by a runtime wrapper that fills the following slots:
\begin{itemize}[leftmargin=*,nosep]
    \item System instruction and task instruction.
    \item Fact Base, containing confirmed information from the Oracle.
    \item Questions asked so far.
    \item Active Spatial Hint, spatial progress since the last step, and Type~4 follow-up state.
    \item Spatial Memory, last-step purpose, movement history, and offset from start.
    \item Current observation image.
    \item Global heading for the current front view.
    \item Steps taken and questions remaining.
\end{itemize}
The final runtime instruction is to output the decision as a single JSON object with no markdown fences.
\end{tcolorbox}

\begin{tcolorbox}[title=Grounder template,colback=gray!4,colframe=black!45,boxrule=0.4pt,arc=1mm,left=4pt,right=4pt,top=4pt,bottom=4pt]
\small
\textbf{Role.} The Grounder is a visual grounding agent. It receives the Planner's subgoal and a single selected $90^\circ$ view. Its only task is to select the grid cell the robot should walk toward. It does not choose the navigation strategy.

\textbf{Semantic grounding first.} The grounding instruction requires three steps in order: parse the navigation instruction semantically, locate the relevant object, doorway, passage, corridor opening, or free-space target in the image, and then choose the grid cell for that located target. It explicitly forbids selecting a default cell without first finding the instruction-relevant target.

\textbf{Grid convention.} The selected view contains an $8\times8$ grid. Columns are numbered 1 to 8 from left to right, with columns 4 and 5 near centre. Rows are numbered from bottom to top. Row 1 is floor level, row 8 is ceiling level, rows 1 and 2 are nearby ground targets, row 3 is for farther targets such as doorways or corridor ends, and rows 4 to 8 are discouraged for navigation because they often hit walls, furniture tops, or ceiling.

\textbf{Counting and safety rules.} Before choosing a row, the Grounder must count strips upward from the bottom edge. For doorways, hallways, and narrow passages, it must select the centre of the opening rather than an edge. For row 3, the prompt requires an open-space check: the target opening must be clearly far, the centre path must look traversable, and a longer move must be desirable. Otherwise row 2 is preferred.

\textbf{Grounder runtime prompt and output.} The runtime prompt is \texttt{\{system\}}, \texttt{Navigation instruction: \{subgoal\}}, \texttt{Current Observation:<image>}, and a request to select one grid cell. The output is strict JSON:
\begin{verbatim}
{
  "col": 3,
  "row": 2,
  "rationale": "The instruction target is the doorway on the left side..."
}
\end{verbatim}
\end{tcolorbox}

\begin{tcolorbox}[title=Oracle interface and prompt template,colback=gray!4,colframe=black!45,boxrule=0.4pt,arc=1mm,left=4pt,right=4pt,top=4pt,bottom=4pt]
\small
\textbf{How the agent asks.} The Planner asks by setting the ask flag, selecting one of the four question types, and writing a natural-language question. The movement direction and subgoal are left empty, so no movement occurs in that step. The controller sends the typed question to the Oracle, records the exchange, charges the locked type-level cost, and inserts the answer into the Fact Base. Type~1 and Type~2 answers are cached to avoid duplicate global questions. Type~3 answers activate a persistent spatial hint. Type~4 answers update candidate-confirmation state; a positive answer either stops immediately or requests exactly one closer move before stopping.

\textbf{Oracle knowledge and restrictions.} The Oracle receives typed access to benchmark metadata, not an unrestricted navigation API. Its internal slots are target category, sanitized target appearance description, goal room or region, current observation when needed, candidate query when Type~4 is asked, and a Type~3 summary derived from front view observations sampled at navigation waypoints. The waypoint views are used only to form a coarse natural-language spatial hint; they are not exposed to the Planner as frames, camera poses, waypoint coordinates, path lengths, or action sequences. The answer must obey the requested type. Type~1 can describe appearance but cannot reveal room or route. Type~2 can name the room or region but cannot add appearance cues. Type~3 can give a natural-language route or layout hint, but cannot output metric coordinates, compass headings, step counts, final target coordinates, exact waypoint sequences, raw video evidence, or exact action sequences. Type~4 only answers whether the visible queried candidate is the target and may add a short rationale. Thus Type~3 emulates a constrained human spatial hint rather than handing the executor a path to follow.

\textbf{Oracle system template.}
\begin{verbatim}
You are a constrained navigation Oracle for an indoor instance-goal task.
Answer only the agent's typed question. Do not volunteer information from
another question type. Keep the answer concise and useful for the next decision.

Question-type rules:
Type 1 appearance: use only the target appearance description.
Type 2 region: use only goal room or region metadata.
Type 3 spatial: use only the waypoint view summary to give coarse
natural-language guidance. Do not reveal raw frames, camera poses, waypoint
coordinates, metric coordinates, compass headings, step counts, path lengths,
the final target coordinate, or an exact action sequence.
Type 4 confirmation: use the current observation and queried candidate.
Answer yes or no, with a short rationale.
\end{verbatim}

\textbf{Oracle runtime template.}
\begin{verbatim}
Target category: {target_category}
Sanitized appearance description: {appearance_description}
Goal room/region: {goal_region}
Waypoint view summary for Type 3: {route_layout_summary}
Current observation for Type 3/4: <image if provided>
Visible candidate for Type 4: {candidate_description}
Prior correction or feedback: {feedback_context}

Agent question (Type {question_type}): {question}
Return a concise answer following the type-specific rules.
\end{verbatim}
\end{tcolorbox}

\subsection{Grounding and Local Control}
After the Planner selects one of the eight directions, the Grounder sees only that direction's $90^\circ$ image. It outputs a column and row on an $8\times8$ grid. Rows are interpreted bottom-up: low rows correspond to ground-level navigation targets, while high rows usually correspond to walls, furniture tops, or ceiling. The Grounder is instructed to select the centre of doorways and passages rather than edges, because edge clicks are the most common cause of local collision or incomplete doorway entry. For the same reason, the Planner's subgoal must name a place rather than an action: ``centre of the open doorway in front-left'' is groundable, whereas ``explore left'' is not.

The executor converts the grounded grid cell into a world-frame waypoint using the calibrated camera model. Let the selected view have width $W$ and height $H$, and let the Grounder output column $c_g$ and bottom-up row $r_g$ in the $8\times8$ grid. We use the centre pixel of that cell,
\begin{equation}
    u_g=\frac{(c_g-\frac{1}{2})W}{8},\qquad
    v_g=H-\frac{(r_g-\frac{1}{2})H}{8}.
    \label{eq:app:grid-to-pixel}
\end{equation}
Given camera intrinsics $K$ and a camera-to-world transform $T_{wc}=[R_{wc}\ \mathbf{t}_{wc};\ \mathbf{0}^{\top}\ 1]$, a valid depth value $z=D(u_g,v_g)$ gives
\begin{equation}
    \mathbf{x}_c=zK^{-1}
    \begin{bmatrix}
        u_g \\ v_g \\ 1
    \end{bmatrix},
    \qquad
    \mathbf{x}_w=R_{wc}\mathbf{x}_c+\mathbf{t}_{wc}.
    \label{eq:app:runtime-camera-world}
\end{equation}
If the simulator or robot stores a world-to-camera transform $T_{cw}$, we first invert it to obtain $T_{wc}$. When the selected pixel has invalid depth, the executor instead intersects the camera ray with the floor plane. With ray direction $\mathbf{d}_w=R_{wc}K^{-1}[u_g,v_g,1]^\top$ and floor plane $\mathbf{n}^{\top}\mathbf{x}+b=0$, the intersection is
\begin{equation}
    \mathbf{x}_w
    =
    \mathbf{t}_{wc}
    -
    \frac{\mathbf{n}^{\top}\mathbf{t}_{wc}+b}
         {\mathbf{n}^{\top}\mathbf{d}_w}
    \mathbf{d}_w .
    \label{eq:app:floor-intersection}
\end{equation}
The waypoint used by the local controller is the floor-plane projection of $\mathbf{x}_w$, clipped to a short horizon around the agent so that a single Grounder output cannot command a long metric jump.

The low-level executor uses raycasting as a lightweight safety layer under the MLLM policy. Before committing to a local waypoint, it casts forward rays at multiple vertical heights so that low furniture, table-level obstacles, and taller occluders can all block motion. If any forward ray detects an obstacle within $0.6$\,m, the executor does not continue straight; it selects the nearest collision-free neighbouring heading and records the blocked direction in Spatial Memory. The rule is local by construction: it prevents short-horizon collisions without replacing the Planner's semantic choice of where to go.

Because the executor maintains both the agent pose and the active navigation target in a shared world coordinate frame, turning around an obstacle does not destroy goal direction information. Let the agent's floor-plane position and heading at step $t$ be $\mathbf{p}_t=(x_t,z_t)$ and $\theta_t$, and let the active target waypoint be $\mathbf{g}=(x_g,z_g)$. The executor recomputes
\begin{equation}
    \Delta_t = \mathbf{g}-\mathbf{p}_t,\quad
    r_t=\|\Delta_t\|_2,\quad
    \beta_t=\mathrm{wrap}\left(\mathrm{atan2}(\Delta_{t,z},\Delta_{t,x})-\theta_t\right),
    \label{eq:app:relative-target}
\end{equation}
after every move or turn. Here $r_t$ is the remaining world-frame distance to the active target and $\beta_t$ is the target's relative bearing in the agent frame. Thus, even if raycasting forces a detour, the controller can continue to report whether the target lies ahead, left, or right of the current heading and can resume progress once the immediate obstacle is cleared.

\section{Computation Resources}
\label{app:compute}

All agent experiments were run in Isaac Sim 4.5 on NVIDIA RTX A6000 GPUs with 48GB of memory. The main cost comes from repeated MLLM calls during interactive evaluation: each episode requires iterative Planner calls, conditional Grounder calls for movement steps, and Oracle calls only when the agent asks a question. Benchmark construction uses additional MLLM calls for reference-view filtering and caption generation as described in Section~\ref{app:benchmark}; these calls are performed offline and are not part of agent evaluation.

All interactive evaluations use deterministic decoding with temperature set to zero, a maximum of 20 Planner steps per episode, a maximum budget of 10 Oracle questions, and a 1024 token generation cap for each Planner, Grounder, and Oracle call. The same decoding settings, prompts, question costs, step limit, and question budget are reused for all reported agent variants and model backbones. Episodes are not rerun with modified prompts or budgets after observing success or failure.

\section{Real-World Deployment}
\label{app:real-world}

The agent design is compatible with real robot deployment because the MLLM policy is separated from the low-level motion layer. A physical platform would provide an RGB-D or multi-camera observation, camera intrinsics, a calibrated camera-to-robot transform, odometry or SLAM pose estimates in a world frame, and a short-range safety sensor such as depth, LiDAR, or ultrasonic ranging. The Planner would still operate on an 8-direction visual observation, Spatial Memory, Fact Base, movement history, and question budget; the Grounder would still select a local grid target in the selected view; and the executor would convert that target into a short local motion primitive subject to robot-specific safety constraints.

In this deployment, the runtime camera-to-world conversion in Section~\ref{app:agent} serves as the interface between perception and control rather than as privileged oracle information. Camera calibration and pose estimation allow the system to express the current robot pose, grounded waypoint, and active target hint in one coordinate frame. The raycasting safety layer can be implemented with depth or LiDAR rays at multiple heights, while the same relative-bearing computation in Eq.~\eqref{eq:app:relative-target} keeps the agent aware of whether the target remains ahead, left, or right after a detour. The oracle interaction can be provided by a human user or by an application-specific knowledge source; in either case, the question type and answer are logged in the Fact Base so that the agent's use of interaction remains auditable.

\section{Limitations}
\label{app:limitations}

Our geometric execution assumes an idealised sensing and localisation interface. In real physical environments, camera intrinsics/extrinsics are noisy, robot poses drift, depth has missing values, and actuation introduces small deviations from the intended waypoint. These perturbations can affect the camera-to-world conversion, the relative-bearing estimate, and the raycasting safety decision. A deployed system would need calibration checks, uncertainty-aware pose filtering, and conservative safety margins around obstacles.

The current MLLM policy also does not consume all navigation signals that a robot could provide. It reasons over discrete views, language memory, question answers, heading, and short movement history, but it does not fully use continuous video, dense temporal motion cues, wheel odometry traces, inertial signals, or a learned local traversability map. As a result, some information that is available during navigation is compressed before reaching the Planner. Future versions could incorporate video-context models, richer temporal memory, or learned controllers that expose more of these signals without overwhelming the language planner.

Finally, the interaction loop improves reliability but is not yet optimised for efficiency. Repeated MLLM calls increase latency, and the ask/move/ground cycle can be slower than a specialised embodied policy. More efficient deployment would require caching, smaller local models for routine grounding, asynchronous perception updates, and adaptive question policies that account for both success probability and real-time cost.

\section{Broader Impact}
\label{app:broader-impact}

Interactive embodied navigation shifts the problem from passive instruction following to situated collaboration. This has positive implications for assistive robotics, household service agents, and inspection systems: when an agent is uncertain, asking a targeted question can be safer and more efficient than silently guessing. The same framing also makes evaluation more realistic, because success depends not only on reaching a location but on using human feedback responsibly.

The interaction channel also introduces risks. An agent that asks poorly calibrated questions can burden users, reveal private information through its observations, or overfit to social cues instead of grounding decisions in the environment. Systems deployed in homes or workplaces must therefore limit what is stored, make question histories auditable, and avoid using interaction as a way to extract unnecessary personal information. Our benchmark focuses on object search in static indoor scenes and does not address all of these deployment issues.

For MLLM agents, the broader methodological impact is that interaction exposes failure modes hidden by one-shot navigation benchmarks. A strong agent must decide when it knows enough, when it should ask, and how to convert language answers into grounded action. This encourages research on calibrated uncertainty, memory, controllable question policies, and transparent agent state. At the same time, stronger MLLM agents may make failures look fluent even when their spatial beliefs are wrong; reporting question costs, memory state, and failure cases is therefore important for responsible comparison.

\section{Use of Large Language Models}
\label{app:llm-use}

We used large language models in two roles. First, as part of the proposed method, MLLMs instantiate the Planner, Grounder, Oracle, uncertainty scorer, and the offline captioning/filtering components described above. These uses are algorithmic and are reported as part of the experimental protocol. Second, for paper preparation, language models were used only for prose polishing, grammar editing, and improving the readability of captions and appendix descriptions. They were not used to fabricate experimental results, change metric values, or decide which runs to report; all reported numbers come from the evaluation records or explicitly marked proxy analyses.